# Order of Magnitude Comparisons of Distance

**Ernest Davis**                                                                 DAVISE@CS.NYU.EDU
*Courant Institute*
*New York, NY 10012 USA*

## Abstract

Order of magnitude reasoning — reasoning by rough comparisons of the sizes of quantities — is often called "back of the envelope calculation", with the implication that the calculations are quick though approximate. This paper exhibits an interesting class of constraint sets in which order of magnitude reasoning is demonstrably fast. Specifically, we present a polynomial-time algorithm that can solve a set of constraints of the form "Points $a$ and $b$ are much closer together than points $c$ and $d$." We prove that this algorithm can be applied if "much closer together" is interpreted either as referring to an infinite difference in scale or as referring to a finite difference in scale, as long as the difference in scale is greater than the number of variables in the constraint set. We also prove that the first-order theory over such constraints is decidable.

## 1. Introduction

Order of magnitude reasoning — reasoning by rough comparisons of the sizes of quantities — is often called "back of the envelope calculation", with the implication that the calculations are quick though approximate. Previous AI work on order of magnitude reasoning, however, has focussed on its expressive power and inferential structure, not on its computational leverage (Raiman, 1990; Mavrovouniotis and Stephanopoulos, 1990; Davis, 1990; Weld, 1990).

In this paper we exhibit an interesting case where solving a set of order of magnitude comparisons is demonstrably much faster than solving the analogous set of simple order comparisons. Specifically, given a set of constraints of the form "Points $a$ and $b$ are much closer together than points $c$ and $d$," the consistency of such a set can be determined in low-order polynomial time. By contrast, it is easily shown that solving a set of constraints of the form "The distance from $a$ to $b$ is less than or equal to the distance from $c$ to $d$" in one dimension is NP-complete, and in higher dimensions is as hard as solving an arbitrary set of algebraic constraints over the reals.

In particular, the paper presents the following results:

1. The algorithm "solve_constraints($\mathcal{S}$)" solves a system of constraints of the form "Points $a$ and $b$ are infinitely closer than points $c$ and $d$" in polynomial time (Section 5).

2. An improved version of the algorithm runs in time $O(\max(n^2\alpha(n)), ne, s)$ where $n$ is the number of variables, $\alpha(n)$ is the inverse Ackermann's function, $e$ is the number of edges mentioned in the constraint set, and $s$ is the size of the constraint set. (Section 6.1).

3. An extended version of the algorithm allows the inclusion of non-strict constraints of the form "Points $a$ and $b$ are not infinitely further apart than points $c$ and $d$." The running time for this modified algorithm is slower than that of solve_constraints, but still polynomial time. (Section 6.2)

4. A different extension of the algorithm allows the combination of order of magnitude constraints on distances with order comparisons on the points of the form "Point $a$ precedes point $b$." (Section 6.3)

5. The same algorithm can be applied to constraints of the form "The distance from $a$ to $b$ is less than $1/B$ times the distance from $c$ to $d$," where $B$ is a given finite value, as long as $B$ is greater than the number of variables in the constraint set. (Section 7)

6. The first-order theory over such constraints is decidable. (Section 8)

As preliminary steps, we begin with a small example and an informal discussion (Section 2). We then give a formal account of order-of-magnitude spaces (Section 3) and present a data structure called a *cluster tree*, which expresses order-of-magnitude distance comparisons (Section 4). We conclude the paper with a discussion of the significance of these results (Section 9).

## 2. Examples

Consider the following inferences:

**Example 1:** I wish to buy a house and rent office space in a suburb of Metropolis. For obvious reasons, I want the house to be close to the school, the house to be close to the office, and the office to be close to the commuter train station. I am told that in Elmville the train station is quite far from the school, but in Newton they are close together.

**Infer** that I will not be able to satisfy my constraints in Elmville, but may be able to in Newton.

**Example 2:** The Empire State Building is much closer to the Washington Monument than to Versailles. The Statue of Liberty is much closer both to the Empire State Building and to Carnegie Hall than to the Washington Monument.

**Infer** that Carnegie Hall is much closer to the Empire State Building than to Versailles.

**Example 3:** You have to carry out a collection of computational tasks covering a wide range of difficulty. For instance

a. Add up a column of 100 numbers.

b. Sort a list of 10,000 elements.

c. Invert a $100 \times 100$ matrix.

d. Invert a $1000 \times 1000$ matrix.

e. Given the O.D.E. $\ddot{x} = \cos(e^t x)$, $x(0) = 0$, find $x(20)$ to 32-bit accuracy.

f. Given a online collection of 1,000 photographs in GIF format, use state-of-the-art image recognition software to select all those that show a man on horseback.

g. Do a Web search until you have collected 100 pictures of men on horseback, using state-of-the-art image recognition software.

h. Using state-of-the-art theorem proving software, find a proof that the medians of a triangle are concurrent.

i. Using state-of-the-art theorem proving software, find a proof of Fermat's little theorem.

It is plausible to suppose that, in many of these cases, you can say reliably that one task will take much longer than another, either by a human judgment or using an expert system. For instance, task (a) is much shorter than any of the others. Task (b) is much shorter than any of the others except (a) and possibly (h). Task (c) is certainly much shorter than (d), (f), (g), or (i). However, with certain pairs such as (c) and (h) or (c) and (e) it would be difficult to guess whether one is much shorter than another, or whether they are of comparable difficulty.

You have a number of independent identical computers, of unknown vintage and characteristics, on which you will schedule tasks of these kinds. Note that, under these circumstances, there is no way to predict the absolute time required by any of these tasks within a couple of orders of magnitude. Nonetheless, the comparative lengths presumably still stand.

**Given:** a particular schedule of tasks on machines, infer what you can about the relative order of completion times. For example, given the following schedule

Machine M1: tasks a,b,h,d.
Machine M2: tasks c,i.

it should be possible to predict that (a) and (b) will complete before (c); that (c) will complete before (d); and that (d) will complete before (i); but it will not be possible to predict the order in which (c) and (h) will complete.

In all three examples, the given information has the form "The distance between points $W$ and $X$ is much less than the distance between $Y$ and $Z$". In examples 1 and 2, the points are geometric. In example 3, the points are the start and completion times of the various tasks, and the constraints on relative lengths can be put in the form "The distance from start(a) to end(b) is much less than the distance from start(c) to end(c)", and so on. In example 3, there is also ordering information: the start of each task precedes its end; the end of (a) is equal to the start of (b); and so on. The problem is to make inferences based on this weak kind of constraint.

It should be noted that these examples are meant to be illustrative, rather than serious applications. Example 1 does not extend in any obvious way to a class of natural, large problems. Example 2 is implausible as a state of knowledge; how does the reasoner find himself knowing just the order-of-magnitude relations among distances and no other geometric information? Example 3 is contrived. Nonetheless, these illustrate the kinds of situations where order-of-magnitude relations on distance do arise; where they express a substantial part of the knowledge of the reasoner; and where inferences based purely on the order-of-magnitude comparisons can yield useful conclusions.

The methods presented in this paper involve construing the relation "Distance D is much shorter than distance E" as if it were "Distance D is infinitesimal as compared to distance E." As we shall see, under this interpretation, systems of constraints over distances can be solved efficiently. The logical foundations for dealing with infinitesimal quantities lie in the non-standard model of the real line with infinitesimals, developed by Abraham Robinson (1965). (A more readable account is given by Keisler, 1976.) Reasoning with quantities of infinitely different scale is known as "order of magnitude" reasoning.

The reader may ask, "Since infinitesimals have no physical reality, what is the value of developing techniques for reasoning about them?" In none of the examples, after all, is the smaller quantity truly infinitesimal or the larger one truly infinite. In example 1 and 2, the ratio between successive sizes is somewhere between 10 and 100; in example 3, it is between 100 and a rather large number difficult to estimate; but one can always give some kind of upper bound. It is essentially certain, for instance, that the ratio between the times required for tasks (a) and (i) is less than $10^{100,000}$. Why not use the best real-valued estimate instead?

The first answer is that this is an idealization. Practically all physical reasoning and calculation rest on one idealization or another: the idealization in the situation calculus that time is discrete; the idealization that solid objects are rigid, employed in most mechanics programs; the idealization that such physical properties as density, temperature, and pressure are continuous rather than local averages over atoms, which underlies most uses of partial differential equations; the idealization involved in the use of the Dirac delta function; and so on. Our idealization here that a very short distance is infinitesimally smaller than a long one simplifies reasoning and yields useful results as long as care is taken to stay within an appropriate range of application.

The second answer is that this is a technique of mathematical approximation, which we are using to turn an intractable problem into a tractable one. This would be analogous to linearizing a non-linear equation over a small neighborhood; or to approximating a sum by an integral.

There are circumstances where we can be sure that the approximation gives an answer that is guaranteed exactly correct; namely if the actual ratio implicit in the comparison "$D$ is much smaller than $E$" is larger than the number of points involved in the system of constraints. This will be proven in Section 7. There is also a broader, less well-defined, class of problems where the approximation, though not guaranteed correct, is more reliable than some of the other links in the reasoning. For instance, suppose that one were to consider an instance of example 3 involving a couple of hundred tasks, apply order-of-magnitude reasoning, and come up with an answer that can be determined to be wrong. It is possible that the error would be due to the order-of-magnitude reasoning. However, it seems safe to say that, in most cases, the error is more likely to be due to a mistake in estimating the comparative sizes.

## 3. Order-of-magnitude spaces

An order-of-magnitude space, or *om-space*, is a space of geometric points. Any two points are separated by a distance. Two distances $d$ and $e$ are compared by the relation $d \ll e$, meaning "Distance $d$ is infinitesimal compared to $e$" or, more loosely, "Distance $d$ is much smaller than $e$."

For example, let $\Re^*$ be the non-standard real line with infinitesimals. Let $\Re^{*m}$ be the corresponding $m$-dimensional space. Then we can let a point of the om-space be a point in $R^{*m}$. The distance between two points $a, b$ is the Euclidean distance, which is a non-negative value in $\Re^*$. The relation $d \ll e$ holds for two distances $d, e$, if $d/e$ is infinitesimal.

The distance operator and the comparator are related by a number of axioms, specified below. The most interesting of these is called the *om-triangle inequality*: If $ab$ and $bc$ are both much smaller than $xy$, then $ac$ is much smaller than $xy$. This combines the ordinary triangle inequality "The distance $ac$ is less than or equal to distance $ab$ plus distance $bc$" together with the rule from order-of-magnitude algebra, "If $p \ll r$ and $q \ll r$ then $p+q \ll r$."

It will simplify the exposition below if, rather than talking about distances, we talk about orders of magnitude. These are defined as follows. We say that two distances $d$ and $e$ have the *same* order of magnitude if neither $d \ll e$ nor $e \ll d$. In $\Re^*$ this is the condition that $d/e$ is finite: neither infinitesimal nor infinite. (Raiman, 1990 uses the notation "$d$ Co $e$" for this relation.) By the rules of the order-of-magnitude calculus, this is an equivalence relation. Hence we can define *an order of magnitude* to be an equivalence class of distances under the relation "same order of magnitude". For two points $a, b$, we define the function $\mathrm{od}(a, b)$ to be the order of magnitude of the distance from $a$ to $b$. For two orders of magnitude $p, q$, we define $p \ll q$ if, for any representatives $d \in p$ and $e \in q$, $d \ll e$. By the rules of the order-of-magnitude calculus, if this holds for any representatives, it holds for all representatives. The advantage of using orders-of-magnitude and the function "od", rather than distances and the distance function, is that it allows us to deal with logical equality rather than the equivalence relation "same order of magnitude".

For example, in the non-standard real line, let $\delta$ be a positive infinitesimal value. Then values such as $\{1, 100, 2 - 50\delta + 100\delta^2 \ldots\}$, are all of the same order of magnitude, $o1$. The values $\{\delta, 1.001\delta, 3\delta + e^{-1/\delta} \ldots\}$ are of a different order of magnitude $o2 \ll o1$. The values $\{1/\delta, 10/\delta + \delta^5 \ldots\}$ are of a third order of magnitude $o3 \gg o1$.

**Definition 1:** An *order-of-magnitude space (om-space)* $\Omega$ consists of:

- A set of points $\mathcal{P}$;

- A set of orders of magnitude $\mathcal{D}$;

- A distinguished value $0 \in \mathcal{D}$;

- A function "$\mathrm{od}(a, b)$" mapping two points $a, b \in \mathcal{P}$ to an order of magnitude;

- A relation "$d \ll e$" over two orders of magnitude $d, e \in \mathcal{D}$

satisfying the following axioms:

A.1 For any orders of magnitude $d, e \in \mathcal{D}$, exactly one of the following holds: $d \ll e$, $e \ll d$, $d = e$.

A.2 For $d, e, f \in \mathcal{D}$, if $d \ll e$ and $e \ll f$ then $d \ll f$.
(Transitivity. Together with A.1, this means that $\ll$ is a total ordering on orders of magnitude.)

A.3 For any $d \in \mathcal{D}$, not $d \ll 0$.
(0 is the minimal order of magnitude.)

A.4 For points $a, b \in \mathcal{P}$, $\text{od}(a, b) = 0$ if and only if $a = b$.
   (The function od is positive definite.)

A.5 For points $a, b \in \mathcal{P}$, $\text{od}(a, b) = \text{od}(b, a)$.
   (The function od is symmetric.)

A.6 For points $a, b, c \in \mathcal{P}$, and order of magnitude $d \in \mathcal{D}$,
   if $\text{od}(a, b) \ll d$ and $\text{od}(b, c) \ll d$ then $\text{od}(a, c) \ll d$.
   (The om-triangle inequality.)

A.7 There are infinitely many different orders of magnitude.

A.8 For any point $a_1 \in \mathcal{P}$ and order of magnitude $d \in \mathcal{D}$, there exists an infinite set
   $a_2, a_3 \ldots$ such that $\text{od}(a_i, a_j) = d$ for all $i \neq j$.

The example we have given above of an om-space, non-standard Euclidean space, is wild and woolly and hard to conceptualize. Here are two simpler examples of om-spaces:

I. Let $\delta$ be an infinitesimal value. We define a point to be a polynomial in $\delta$ with integer coefficients, such as $3 + 5\delta - 8\delta^5$. We define an order-of-magnitude to be a power of $\delta$. We define $\delta^m \ll \delta^n$ if $m > n$; for example, $\delta^6 \ll \delta^4$. We define $\text{od}(a, b)$ to be the smallest power of $\delta$ in $a - b$. For example, $\text{od}(1 + \delta^2 - 3\delta^3, 1 - 5\delta^2 + 4\delta^4) = \delta^2$.

II. Let $N$ be an infinite value. We define a point to be a polynomial in $N$ with integer coefficients. We define an order of magnitude to be a power of $N$. We define $N^p \ll N^q$ if $p < q$; for example, $N^4 \ll N^6$. We define $\text{od}(a, b)$ to be the largest power of $N$ in $a - b$. For example, $\text{od}(1 + N^2 - 3N^3, 1 - 5N^2 + 4N^4) = N^4$.

It can be shown that any om-space either contains a subset isomorphic to (I) or a subset isomorphic to (II). (This is just a special case of the general rule that any infinite total ordering contains either an infinite descending chain or an infinite ascending chain.)

We will use the notation "$d \underset{\approx}{\ll} e$" as an abbreviation for "$d \ll e$ or $d = e$".

## 4. Cluster Trees

Let $P$ be a finite set of points in an om-space. If the distances between different pairs of points in $P$ are of different orders of magnitude, then the om-space imposes a unique tree-like hierarchical structure on $P$. The points will naturally fall into *clusters,* each cluster $C$ being a collection of points all of which are much closer to one another than to any point in $P$ outside $C$. The collection of all the clusters over $P$ forms a strict tree under the subset relation. Moreover, the structure of this tree and the comparative sizes of different clusters in the tree captures all of the order-of-magnitude relations between any pair of points in $P$. The tree of clusters is thus a very powerful data structure for reasoning about points in an om-space, and it is, indeed, the central data structure for the algorithms we will develop in this paper. In this section, we give a formal definition of cluster trees and prove some basic results as foundations for our algorithms.

**Definition 2:** Let $P$ be a finite set of points in an om-space. A non-empty subset $C \subset P$ is called a *cluster* of $P$ if for every $x, y \in C$, $z \in P - C$, $\text{od}(x, y) \ll \text{od}(x, z)$. If $C$ is a cluster, the diameter of $C$, denoted "odiam$(C)$", is the maximum value of $\text{od}(x, y)$ for $x, y \in C$.

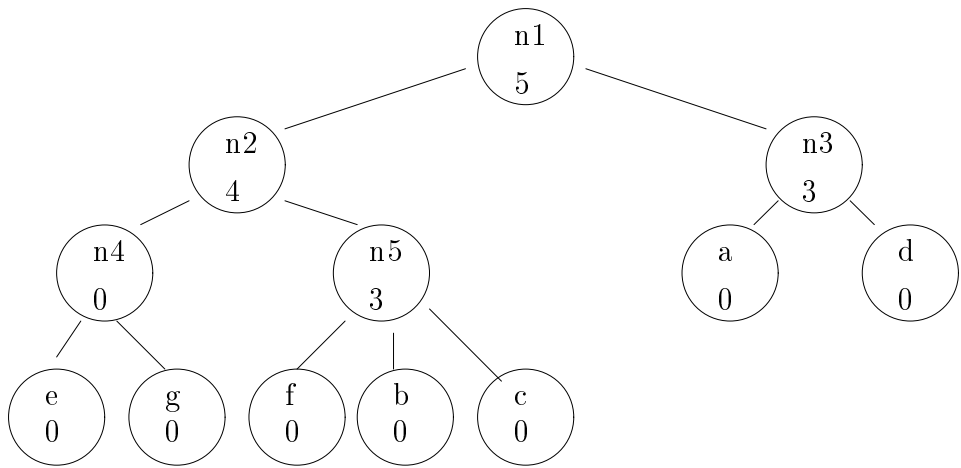

Figure 1: Cluster tree

Note that the set of any single element of $P$ is trivially a cluster of $P$. The entire set $P$ is likewise a cluster of $P$. The empty set is by definition not a cluster of $P$.

**Lemma 1:** If $C$ and $D$ are clusters of $P$, then either $C \subset D$, $D \subset C$, or $C$ and $D$ are disjoint.

**Proof:** Suppose not. Then let $x \in C \cap D$, $y \in C - D$, $z \in D - C$. Since $C$ is a cluster, $\mathrm{od}(x,y) \ll \mathrm{od}(x,z)$. Since $D$ is a cluster, $\mathrm{od}(x,z) \ll \mathrm{od}(x,y)$. Thus we have a contradiction. □

By virtue of lemma 1, the clusters of a set $P$ form a tree. We now develop a representation of the order of magnitude relations in $P$ by constructing a tree whose nodes correspond to the clusters of $P$, labelled with an indication of the relative size of each cluster.

**Definition 3:** A *cluster tree* is a tree $T$ such that

- Every leaf of $T$ is a distinct symbol.

- Every internal node of $T$ has at least two children.

- Each internal node of $T$ is labelled with a non-negative value. Two or more nodes may be given the same value. (For the purposes of Sections 5-7, labels may be taken to be non-negative integers; in Section 8, it will be useful to allow rational labels.)

- Every leaf of the tree is labelled 0.

- The label of every internal node in the tree is less than the label of its parent.

For any node $N$ of $T$, the field "$N$.symbols" gives the set of symbols in the leaves in the subtree of $T$ rooted at $N$, and the field "$N$.label" gives the integer label on node $N$.

Thus, for example, in Figure 1, n3.label=3 and n3.symbols = $\{a, d\}$; n1.label = 5 and n1.symbols = $\{a, b, c, d, e, f, g\}$.

As we shall see, the nodes of the tree $T$ represent the clusters of a set of points, and the labels represent the relative sizes of the diameters of the clusters.

**Definition 4:** A *valuation* over a set of symbols is a function mapping each symbol to a point in an om-space. If $T$ is a cluster tree, a valuation over $T$ is a valuation over $T$.symbols. If $N$ is any node in $T$ and $\Gamma$ is a valuation over $T$, we will write $\Gamma(N)$ as an abbreviation for $\Gamma(N.\text{symbols})$.

We now define how a cluster tree $T$ expresses the order of magnitude relations over a set of points $P$.

**Definition 5:** Let $T$ be a cluster tree and let $\Gamma$ be a valuation over $T$. Let $P = \Gamma(T)$, the set of points in the image of $T$ under $\Gamma$. We say that $\Gamma \models T$ (read $\Gamma$ *satisfies* or *instantiates* $T$) if the following conditions hold:

  i. For any internal node $N$ of $T$, $\Gamma(N)$ is a cluster of $P$.

  ii. For any cluster $C$ of $P$, there is a node $N$ such that $C = \Gamma(N)$.

  iii. For any nodes $M$ and $N$, if $M.\text{label} < N.\text{label}$ then odiam$(\Gamma(M)) \ll$ odiam$(\Gamma(N))$.

  iv. If label$(M) = 0$, then odiam$(M) = 0$. (That is, all children of $M$ are assigned the same value under $\Gamma$.)

The following algorithm generates an instantiation $\Gamma$ given a cluster tree $T$:

**procedure** instantiate(**in** $T$ : cluster tree; $\Omega$ : an om-space)
               **return** : array of points indexed on the symbols of $T$

**variable** $G[N]$ : array of points indexed on the nodes of $T$;

Let $k$ be the number of internal nodes in $T$;
Choose $\delta_0 = 0 \ll \delta_1 \ll \delta_2 \ll \ldots \ll \delta_k$ to be $k + 1$ different orders of magnitude;
            /* Such values can be chosen by virtue of axiom A.7 */
pick a point $x \in \Omega$;
$G[\text{root of } T] := x$;
instantiate1$(T, \Omega, \delta_1 \ldots \delta_k, G)$;
**return** the restriction of $G$ to the symbols of $T$.
**end** instantiate.

instantiate1(**in** $N$ : a node in a cluster tree; $\Omega$ : an om-space; $\delta_1 \ldots \delta_k$ : orders of magnitude;
      **in out** G : array of points indexed on the nodes of $T$)
**if** $N$ is not a leaf **then**
     let $C_1 \ldots C_p$ be the children of $N$;
     $\mathbf{x}_1 := G[N]$;
     $q := N.\text{label}$;
     pick points $x_2 \ldots x_p$ such that
          for all $i, j \in 1 \ldots p$, if $i \neq j$ then od$(x_i, x_j) = \delta_q$;
          /* Such points can be chosen by virtue of axiom A.8 */

```
        for i = 1 . . . p do
            G[C_i] := x_i;
            instantiate1(C_i, Ω, δ_1 . . . δ_k, G);
        endfor
endif end instantiate1.
```

Thus, we begin by picking orders of magnitude corresponding to the values of the labels. We pick an arbitrary point for the root of the tree, and then recurse down the nodes of the tree. For each node $N$, we place the children at points that all lie separated by the desired diameter of $N$. The final placement of the leaves is then the desired instantiation.

**Lemma 2:** If $T$ is a cluster tree and $Ω$ is an om-space, then instantiate($T, Ω$) returns an instantiation of $T$.

The proof is given in the appendix.

Moreover, it is clear that any instantiation $Γ$ of $T$ can be generated as a possible output of instantiate($T, Ω$). (Given an instantiation $Γ$, just pick $G[N]$ at each stage to be $Γ$ of some symbol of $N$.)

Note that, given any valuation $Γ$ over a finite set of symbols $S$, there exists a cluster tree $T$ such that $T$.symbols $= S$ and $Γ$ satisfies $T$. Such a $T$ is essentially unique up to an isomorphism over the set of labels that preserves the label 0 and the order of labels.

## 5. Constraints

In this section, we develop the first of our algorithms. Algorithm solve_constraints tests a collection of constraints of the form "$a$ is much closer to $b$ than $c$ is to $d$," for consistency. If the set is consistent, then the algorithm returns a cluster tree that satisfies the constraints. The algorithm builds the cluster tree from top to bottom dealing first with the large distances, and then proceeding to smaller and smaller distances.

Let $\mathcal{S}$ be a system of constraints of the form od($a, b$) $\ll$ od($c, d$); and let $T$ be a cluster tree. We will say that $T \vdash \mathcal{S}$ (read "$T$ satisfies $\mathcal{S}$") if every instantiation of $T$ satisfies $\mathcal{S}$. In this section, we develop an algorithm for finding a cluster tree that satisfies a given set of constraints.

The algorithm works along the following lines: Suppose we have a solution satisfying $\mathcal{S}$. Let $D$ be the diameter of the solution. If $\mathcal{S}$ contains a constraint od($a, b$) $\ll$ od($c, d$) then, since od($c, d$) is certainly no more than $D$, it follows that od($a, b$) is much smaller than $D$. We label $ab$ as a "short" edge.

If two points $u$ and $v$ are connected by a path of short edges, then by the triangle inequality the edge $uv$ is also short (i.e. much shorter than $D$). Thus, if we compute the connected components $H$ of all the edges that have been labelled short, then all these edges in $H$ can likewise be labelled short. For example, in table 3, edges $vz$, $wx$, and $xy$ can all be labelled "short".

On the other hand, as we shall prove below, if an edge is not in the set $H$, then there is no reason to believe that it is much shorter than $D$. We can, in fact, safely posit that it is the same o.m. as $D$. We label all such edges "long".

We can now assume that any connected component of points connected by short edges is a cluster, and a child of the root of the cluster tree. The root of the cluster tree is then given the largest label. Its children will be given smaller labels. Each "long" edge now connects symbols in two different children of the root. Hence, any instantiation of the tree will make any long edge longer than any short edge.

If no edges are labelled "long" — that is, if $H$ contains the complete graph over the symbols — then there is an inconsistency; all edges are much shorter than the longest edge. For instance, in table 4, since $vw$, $wx$, and $xy$ are all much smaller than $zy$, it follows by the triangle inequality that $vy$ is much smaller than $zy$. But since we also have the constraints that $zy$ is much smaller than $vz$ and that $vz$ is much smaller than $vy$, we have an inconsistency.

The algorithm then iterates, at the next smaller scale. Since we have now taken care of all the constraints $\mathrm{od}(a, b) \ll \mathrm{od}(c, d)$, where $cd$ was labelled "long", we can drop all those from $\mathcal{S}$. Let $D$ now be the greatest length of all the edges that remain in $\mathcal{S}$. If a constraint $\mathrm{od}(a, b) \ll \mathrm{od}(c, d)$ is in the new $\mathcal{S}$, then we know that $\mathrm{od}(a, b)$ is much shorter than $D$, and we label it "short". We continue as above. The algorithm halts when all the constraints in $\mathcal{S}$ have been satisfied, and $\mathcal{S}$ is therefore empty; or when we encounter a contradiction, as above.

We now give the formal statement of this algorithm. The algorithm uses an undirected graph over the variable symbols in $\mathcal{S}$. Given such a graph $G$, and a constraint $C$ of the form $\mathrm{od}(a, b) \ll \mathrm{od}(c, d)$, we will refer to the edge $ab$ as the "short" of $C$, and to the edge $cd$ as the "long" of $C$. The shorts of the system $\mathcal{S}$ is the set of all shorts of the constraints of $\mathcal{S}$ and the longs of $\mathcal{S}$ is the set of all the longs of the constraints. An edge may be both a short and a long of $\mathcal{S}$ if it appears on one side in one constraint and on the other in another constraint.

**procedure** solve_constraints(**in** $\mathcal{S}$: a system of constraints of the form $\mathrm{od}(a, b) \ll \mathrm{od}(c, d)$)
        **return either** a cluster tree $T$ satisfying $\mathcal{S}$ if $\mathcal{S}$ is consistent;
          **or false** if $\mathcal{S}$ is inconsistent.

**type:** A node $N$ of the cluster tree contains
      pointers to the parent and children of $N$;
      the field N.label, holding the integer label;
      and the field N.symbols, holding the list of symbols in the leaves of $N$.

**variables:** $m$ is an integer;
    $C$ is a constraint in $\mathcal{S}$;
    $H, I$ are undirected graphs;
    $N, M$ are nodes of $T$;

**begin if** $\mathcal{S}$ contains any constraint of the form, "$\mathrm{od}(a, b) \ll \mathrm{od}(c, c)$" **then return false**;
   $m :=$ the number of variables in $\mathcal{S}$;
   initialize $T$ to consist of a single node $N$;
   $N$.symbols:= the variables in $\mathcal{S}$;

   **repeat** $H :=$ the connected components of the shorts of $\mathcal{S}$;
      **if** $H$ contains all the edges in $\mathcal{S}$ **then return(false) endif**;
      **for** each leaf $N$ of $T$ **do**
        **if not** all vertices of $N$ are connected in $H$ **then**
          $N$.label := $m$;
          **for** each connected component $I$ of $N$.symbols in $H$ **do**

> construct node $M$ as a new child of $N$ in $T$;
> $M$.symbols:= the vertices of $I$;
> **endfor endif endfor**
> $\mathcal{S}$ := the subset of constraints in $\mathcal{S}$ whose long is in $H$;
> $m := m - 1$;
> **until** $\mathcal{S}$ is empty;
>
> **for** each leaf $N$ of $T$
>     $N$.label := 0;
>     **if** $N$.symbols has more than one symbol
>         **then** create a leaf of $N$ for each symbol in $N$.symbols;
>             label each such leaf 0;
>         **endif endfor end** solve_constraints.

Tables 3 and 4 give two examples of the working of procedure solve_constraints. Table 3 shows how the procedure can be used to establish that the following constraints are consistent:

> The Empire State Building ($x$) is much closer to the Washington Monument ($w$) than to Notre Dame Cathedral ($v$).
> Bunker Hill ($y$) is much closer to the Empire State Building than to the Eiffel Tower ($z$).
> The distance from the Eiffel Tower to Notre Dame is much less than the distance from the Washington Monument to Bunker Hill.

Table 4 shows that the following inference can be justified:

> **Given:** The distances from the Statue of Liberty ($v$) to the World Trade Center ($w$), from the World Trade Center to the Empire State Building ($x$), and from the Empire State Building to the Chrysler Building ($y$) are all much less than the distance from the Chrysler Building to the Washington Monument ($z$).

> **Infer:** The Washington Monument is not much nearer to the Chrysler Building than to the Statue of Liberty.

This inference is carried out by asserting the negation of the consequent, "The Washington Monument is much nearer to the Chrysler Building than to the Statue of Liberty," and showing that that collection of constraints is inconsistent. Note that if we change "much less" and "much nearer" in this example to "less" and "nearer", then the inference no longer valid.

Theorem 1 states the correctness of algorithm solve_constraints. The proof is given in the appendix.

**Theorem 1:** The algorithm solve_constraints($\mathcal{S}$) returns a cluster tree satisfying $\mathcal{S}$ if $\mathcal{S}$ is consistent, and returns **false** if $\mathcal{S}$ is inconsistent.

There may be many cluster trees that satisfy a given set of constraints. Among these, the cluster tree returned by the algorithm solve_constraints has an important property: it has the fewest possible labels consistent with the constraints. In other words, it uses the minimum number of different orders of magnitude of any solution. Therefore, the algorithm can be used to check the satisfiability of a set of constraints in an om-space that violates

$\mathcal{S}$ contains the constraints

1. $\mathrm{od}(w, x) \ll \mathrm{od}(x, v)$.
2. $\mathrm{od}(x, y) \ll \mathrm{od}(y, z)$.
3. $\mathrm{od}(v, z) \ll \mathrm{od}(w, y)$.

The algorithm proceeds as follows:

Initialization:

    The tree is initializes to a single node with $n1$.

    $n1$.symbols := { $v, w, x, y, z$ }.

First iteration:

    The shorts of $\mathcal{S}$ are { $wx, xy, vz$ }.

    Computing the connected components, $H$ is set to { $wx, xy, wy, vz$ }.

    $n1$.label := 5;

    Two children of $n1$ are created:

        $n11$.symbols := $w, x, y$;

        $n12$.symbols := $v, z$;

    As $xv$ is not in $H$, delete constraint #1 from $\mathcal{S}$.

    As $yz$ is not in $H$, delete constraint #2 from $\mathcal{S}$.

    $\mathcal{S}$ now contains just constraint #3.

Second iteration:

    The shorts of $\mathcal{S}$ are { $vz$ }.

    The connected components $H$ is just $\{vz\}$.

    $n11$.label := 4;

    Three children of $n11$ are created:

        $n111$.symbols := $w$;

        $n112$.symbols := $x$;

        $n113$.symbols := $z$;

    As $wy$ is not in $H$, delete constraint #3 from $\mathcal{S}$.

    $\mathcal{S}$ is now empty.

Cleanup:

    $n12$.label := 0;

    Two children of $n12$ are created:

        $n121$.symbols := $v$;

        $n122$.symbols := $z$;

(See Figure 2.)

Table 1: Example of computing a cluster tree

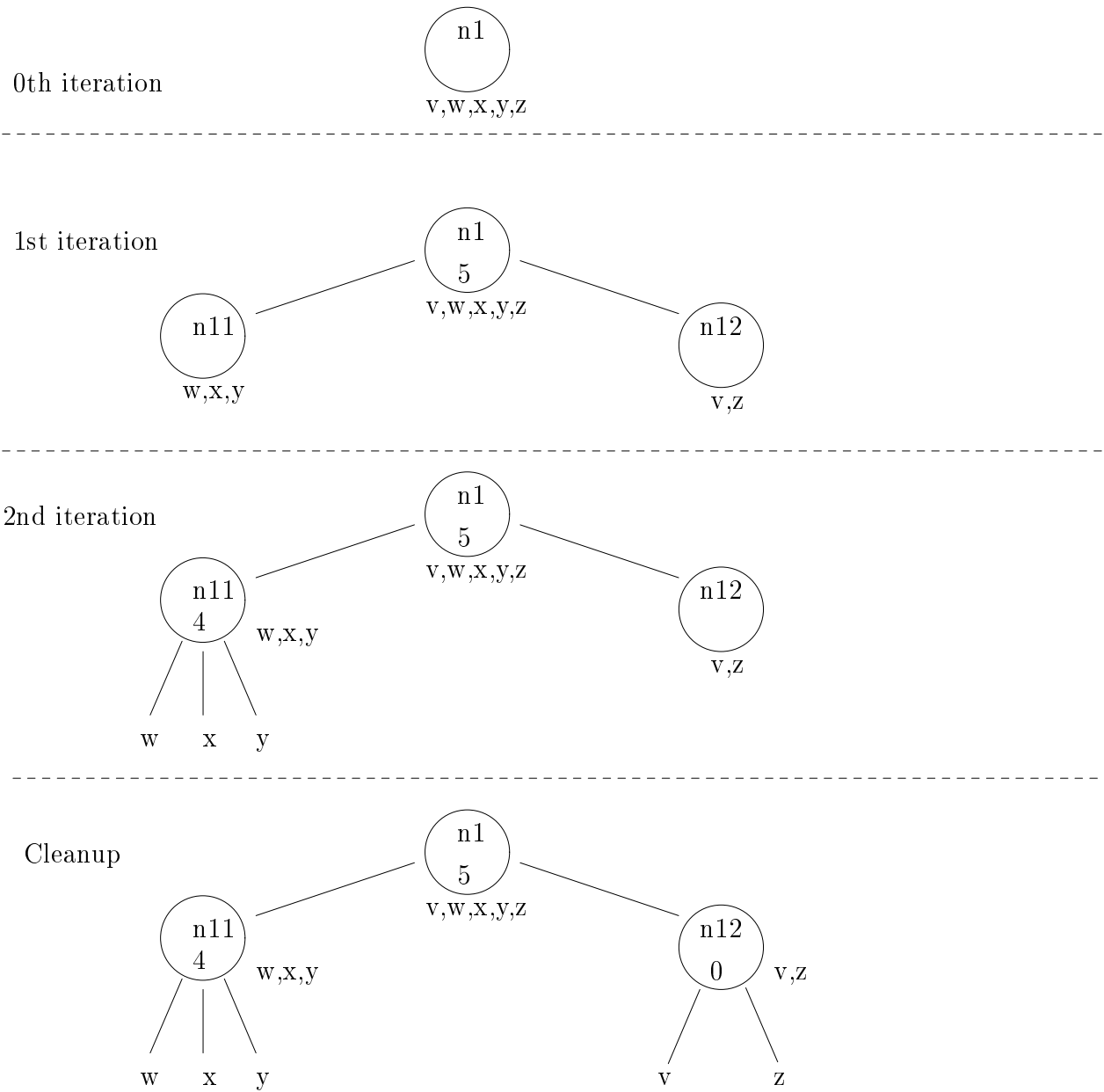

Figure 2: Building a cluster tree

$\mathcal{S}$ contains the constraints

$$\mathrm{od}(v,w) \ll \mathrm{od}(z,y).$$
$$\mathrm{od}(w,x) \ll \mathrm{od}(z,y).$$
$$\mathrm{od}(x,y) \ll \mathrm{od}(z,y).$$
$$\mathrm{od}(z,y) \ll \mathrm{od}(v,z).$$

The algorithm proceeds as follows:

Initialization:

The tree is initializes to a single node with $n1$.

$n1.\mathrm{symbols} := \{\ v,w,x,y,z\ \}.$

First iteration:

The shorts of $\mathcal{S}$ are $\{\ vw, wx, xy, zy, vz\ \}$.

$H$ is set to its connected components, which is the complete graph over $v,w,x,y,z$.

The algorithm exits returning **false**

Table 2: Example of determining inconsistency

axiom A.7 and has only finitely many different orders of magnitude. If the algorithm returns $T$ and $T$ has no more different labels than the number of different orders of magnitude in the space, then the constraints are satisfiable. If $T$ uses more labels than the space has orders of magnitude, then the constraints are unsatisfiable.

The proof is easier to present if we rewrite algorithm solve_constraints in the following form, which returns only the number of different non-zero labels used, but does not actually construct the cluster tree.[1]

**function** num_labels($\mathcal{S}$);
**if** $\mathcal{S}$ is empty **then return**(0)
**else return**(1 + num_labels(reduce_constraints($\mathcal{S}$)))

**function** reduce_constraints($\mathcal{S}$)
$H :=$ connected components of the shorts of $\mathcal{S}$;
**if** $H$ contains all the edges in $\mathcal{S}$ **then return(false)** to top-level
**else return**(the set of constraints in $\mathcal{S}$ whose long is in $H$)

It is easily verified that the sequence of values of $\mathcal{S}$ in successive recursive calls to num_labels is the same as the sequence of values of $\mathcal{S}$ in the main loop of solve_constraints. Therefore num_labels returns the number of different non-zero labels in the tree constructed by solve_constraints.

---

1. The reader may wonder why this simpler algorithm was not presented before the more complicated algorithm solve_constraints. The reason is that the only proof we have found that the system of constraints is consistent if num_labels does not return **false** relies on the relation between num_labels and the constructive solve_constraints.

**Theorem 2:** Out of all solutions to the set of constraints $\mathcal{S}$, the instantiations of solve_constraints($\mathcal{S}$) have the fewest number of different values of od($a, b$), where $a, b$ range over the symbols in $\mathcal{S}$. This number is given by num_labels($\mathcal{S}$).

The proof is given in the appendix.

## 6. Extensions and Consequences

We next present a number of modifications of the algorithm solve_constraints. The first is a more efficient implementation. The second extends the algorithm to handle non-strict comparisons. The third extend the algorithm to handle a combination of order-of-magnitude comparisons on distance with order comparisons, in a one-dimensional space.

### 6.1 An Efficient Implementation of Solve_constraints

It is possible to implement algorithm solve_constraints somewhat more efficiently than the naive encoding of the above description. The key is to observe that the graph $H$ of connected components does not have to be computed explicitly; it suffices to compute it implicitly using merge-find sets (union-find sets). Combining this with suitable back pointers from edges to constraints, we can formulate a more efficient version of the algorithm.

We use the following data structures and subroutines:

- Each node $N$ of the cluster tree contains pointers to its parents and children; a field $N$.label, holding the integer label; a field $N$.symbols, holding the list of symbols in the leaves of $N$; and a field $N$.mfsets, holding a list of the connected components of the symbols in $N$. As described below, each connected component is implemented as an merge-find set (MFSET).

- An edge $E$ in the graph over symbols contains its two endpoints, each of which is a symbol; a field $E$.shorts, a list of the constraints in which $E$ appears as a short; and a field $E$.longs, a list of the constraints in which $E$ appears as a long.

- A constraint $C$ has two fields, $C$.short and $C$.long, both of them edges. It also has pointers into the lists $C$.short.shorts and $C$.long.longs, enabling $C$ to be removed in constant time from the constraint lists associated with the individual edges.

- We will use the disjoint-set forest implementation of MFSETs (Cormen, Leiserson, and Rivest, 1990, p. 448) with merging smaller sets into larger and path-compression. Thus, each MFSET is a upward-pointing tree of symbols, each node of the tree being a symbol. The tree as a whole is represented by the symbol at the root. A symbol $A$ has then the following fields:

  - $A$.parent is a pointer to the parent in the MFSET tree.

  - $A$.cluster_leaf is a pointer to the leaf in the cluster tree containing $A$.

  - If $A$ is the root of the MFSET then $A$.size holds the size of the MFSET.

  - If $A$ is the root of the MFSET, then $A$.symbols holds all the elements of the MFSET.

— If $A$ is the root of the MFSET then $A$.leaf_ptr holds a pointer to the pointer to $A$ in $N$.mfsets where $N = A$.cluster_leaf.

We can now describe the algorithm.

**procedure** solve_constraints1(**in** $\mathcal{S}$: a system of constraints of the form od$(a, b) \ll$ od$(c, d)$).
$\qquad\qquad\qquad\qquad\qquad$ **return either** a cluster tree $T$ satisfying $\mathcal{S}$ if $\mathcal{S}$ is consistent;
$\qquad\qquad\qquad\qquad\qquad\qquad$ **or false** if $\mathcal{S}$ is inconsistent.

**variables:** $m$ is an integer;
$\qquad\qquad$ $a, b$ are symbols;
$\qquad\qquad$ $C$ is a constraint in $\mathcal{S}$;
$\qquad\qquad$ $H$ is an undirected graph;
$\qquad\qquad$ $E, F$ are edges;
$\qquad\qquad$ $P$ is an MFSET;
$\qquad\qquad$ $N, M$ are nodes of $T$;

0. **begin if** $\mathcal{S}$ contains any constraint of the form, "od$(a, b) \ll$ od$(c, c)$" **then return false**;
1. $\qquad$ $H := \emptyset$;
2. $\qquad$ **for** each constraint $C$ in $\mathcal{S}$ with short $E$ and long $F$ **do**
3. $\qquad\qquad$ add $E$ and $F$ to $H$;
4. $\qquad\qquad$ add $C$ to $E$.shorts and to $F$.longs **endfor**;
5. $\qquad$ $m :=$ the number of variables in $\mathcal{S}$;
6. $\qquad$ initialize $T$ to contain the root $N$;
7. $\qquad$ $N$.symbols := the variables in $\mathcal{S}$;

8. $\qquad$ **repeat for** each leaf $N$ of $T$, INITIALIZE_MFSETS($N$);
9. $\qquad\qquad$ **for** each edge $E = ab$ in $H$ **do**
10. $\qquad\qquad\qquad$ **if** $E$.shorts is non-empty and FIND$(a) \neq$ FIND$(b)$ **then**
11. $\qquad\qquad\qquad\qquad$ MERGE(FIND$(a)$, FIND$(b)$) **endif endfor**
12. $\qquad\qquad$ **if every** edge $E = ab$ in $H$ satisfies FIND$(a) =$ FIND$(b)$
13. $\qquad\qquad\qquad$ **then return(false) endif**
14. $\qquad\qquad$ **for** each current leaf $N$ of $T$ do
15. $\qquad\qquad\qquad$ **if** $N$.mfsets has more than one element **then**
16. $\qquad\qquad\qquad\qquad$ **for** each mfset $P$ in $N$.mfsets **do**
17. $\qquad\qquad\qquad\qquad\qquad$ construct node $M$ as a new child of $N$ in $T$;
18. $\qquad\qquad\qquad\qquad\qquad$ $M$.symbols:= $P$.symbols;
19. $\qquad\qquad\qquad\qquad$ **endfor endif endfor**
20. $\qquad\qquad$ **for** each edge $E = ab$ in $H$ **do**
21. $\qquad\qquad\qquad$ **if** FIND$(a) \neq$ FIND$(b)$ **then**
22. $\qquad\qquad\qquad\qquad$ **for** each constraint $C$ in $E$.longs **do**
23. $\qquad\qquad\qquad\qquad\qquad$ delete $C$ from $\mathcal{S}$;
24. $\qquad\qquad\qquad\qquad\qquad$ delete $C$ from $E$.longs;
25. $\qquad\qquad\qquad\qquad\qquad$ delete $C$ from $C$.short.shorts **endfor**
26. $\qquad\qquad\qquad\qquad$ delete $E$ from $H$ **endif endfor**
27. $\qquad\qquad$ $m := m - 1$;
28. $\qquad\qquad$ **until** $\mathcal{S}$ is empty;

29. $\qquad$ **for** each leaf $N$ of $T$
30. $\qquad\qquad$ $N$.label := 0;
31. $\qquad\qquad$ **if** $N$.symbols has more than one symbol

32.          **then** create a leaf of $N$ with label 0 for each symbol in $N$.symbols;
33.          **endif endfor end** solve_constraints1.

**procedure** INITIALIZE_MFSETS($N$ : node)
**var** $A$ : symbol;
$N$.mfsets := $\emptyset$;
**for** $A$ in $N$.symbols **do**
    $A$.parent := null;
    $A$.cluster_leaf := $N$;
    $A$.symbols := $\{A\}$;
    $A$.size := 1;
    $N$.mfsets := cons($A$,$N$.mfsets);
    $A$.leaf_ptr := $N$.mfsets;
**endfor end** INITIALIZE_MFSETS.

**procedure** MERGE(**in** $A, B$ : symbol)
**if** $A$.size $>$ $B$.size **then** swap($A, B$);
$A$.parent := $B$;
$B$.size := $B$.size + $A$.size;
$B$.symbols := $B$.symbols $\cup$ $A$.symbols;
Using $A$.leaf_ptr, delete $A$ from $A$.cluster_leaf.mfsets;
**end** MERGE.

**procedure** FIND(**in** $A$ : symbol) **return** symbol;
**var** $R$ : symbol;
**if** $A$.parent = null **then return** $A$
    **else** $R$ := FIND($A$.parent);
        $A$.parent := $R$; /* Path compression */
        **return**($R$)
**end** FIND.

Let $n$ be the number of symbols in $\mathcal{S}$; let $e$ be the number of edges; and let $s$ be the number of constraints. Note that $n/2 \leq e \leq n(n-1)/2$ and that $e/2 \leq s \leq e(e-1)/2$. The running time of solve_constraints1 can be computed as follows. As each iteration of the main loop 8-28 splits at least one of the connected components of $H$, there can be at most $n-1$ iterations. The MERGE-FIND operations in the **for** loop 9-11 take together time at most $O(\max(n\alpha(n), e))$ where $\alpha(n)$ is the inverse Ackermann's function. Each iteration of the inner **for** loop lines 16-18 creates one node $M$ of the tree. Therefore, there are only $O(n)$ iterations of this loop over the entire algorithm. Lines 14, 15 of the outer **for** loop require at most $n$ iterations in each iteration of the main loop. The **for** loop 22-26 is executed exactly once in the course of the entire execution of the algorithm for each constraint $C$, and hence takes at most time $O(s)$ over the entire algorithm. Steps 20-21 require time $O(e)$ in each iteration of the main loop. It is easily verified that the remaining operations in the algorithm take no more time than these. Hence the overall running time is $O(\max(n^2\alpha(n), ne, s))$.

## 6.2 Adding Non-strict Comparisons

The algorithm solve_constraints can be modified to deal with non-strict comparisons of the form $\text{od}(a, b) \lessapprox \text{od}(c, d)$ by, intuitively, marking the edge $ab$ as "short" on each iteration if the edge $cd$ has been found to be short.

Specifically, in algorithm solve_constraints, we make the following two changes. First, the revised algorithm takes two parameters: $\mathcal{S}$, the set of strict constraints, and $\mathcal{W}$, the set of non-strict constraints. Second, we replace the line

$\quad$ $H :=$ the connected components of the shorts of $\mathcal{S}$

with the following code:

1. $\quad H :=$ the shorts of $\mathcal{S}$;
2. $\quad$ **repeat** $H :=$ the connected components of $H$;
3. $\qquad$ **for** each weak constraint $\text{od}(a, b) \lessapprox \text{od}(c, d)$
4. $\qquad\quad$ **if** $cd$ is in $H$ then add $ab$ to $H$ **endif endfor**
5. $\quad$ **until** no change has been made to $H$ in the last iteration.

The proof that the revised algorithm is correct is only a slight extension of the proof of theorem 1 and is given in the appendix.

Optimizing this algorithm for efficiency is a little involved, not only because of the new operations that must be included, but also because there are now four parameters — $n$, the number of symbols; $e$, the number of edges mentioned; $s$, the number of strict comparison; and $w$, the number of non-strict comparisons — and the optimal implementation varies depending on their relative sizes. In particular, either $s$ or $w$, though not both, may be much smaller than $n$, and each of these cases requires special treatment for optimal efficiency. The best implementation we have found for the case where both $s$ and $w$ are $\Omega(n)$ has a running time of $\text{O}(\max(n^3, nw, s))$. The details of the implementation are straightforward and not of sufficient interest to be worth elaborating here.

An immediate consequence of this result is that a couple of problems of inference are easily computed:

- To determine whether a constraint $C$ is the consequence of a set of constraints $\mathcal{S}$, form the set $\mathcal{S} \cup \neg C$ and check for consistency. If $\mathcal{S} \cup \neg C$ is inconsistent then $\mathcal{S} \models C$. Note that the negation of the constraint $\text{od}(a, b) \ll \text{od}(c, d)$ is the constraint $\text{od}(c, d) \lessapprox \text{od}(a, b)$.

- To determine whether two sets of constraints are logically equivalent, check that each constraint in the first is a consequence of the second, and vice versa.

## 6.3 Adding Order Constraints

Example 3 of Section 2 involves a combination of order-of-magnitude constraints on distances together with simple ordering on points, where the points lie on a one-dimensional line. We next show how to extend algorithm solve_constraints to deal with this more complex situation.

In terms of the axiomatics, adding an ordering on points involves positing that the relation $p < q$ is a total ordering and that the ordering of points is related to order of magnitude comparisons of distances through the following axiom.

A.9 For points $a, b, c \in \mathcal{P}$, if $a < b < c$ then $\mathrm{od}(a, b) \lll \mathrm{od}(a, c)$.

The following rule is easily deduced: If $C$ and $D$ are disjoint clusters, then either every point in $C$ is less than all the points in $D$, or vice versa.

In extending our algorithm, we begin by defining an *ordered cluster tree* to be a cluster tree where, for every internal node $N$, there is a partial order on the children of $N$. If $A$ and $B$ are children of $N$ and $A$ is ordered before $B$, then in an instantiation of the tree, every leaf of $A$ must precede every leaf of $B$. Procedure instantiate1 can then be modified to deal with ordered cluster trees as follows:

instantiate1($\mathbf{in}\ N$ : a node in a cluster tree; $\Omega$ : an om-space; $\delta_1 \ldots \delta_k$ : orders of magnitude;
        $\mathbf{in\ out}$ G : array of points indexed on the nodes of $T$)
$\mathbf{if}$ $N$ is not a leaf $\mathbf{then}$
    let $C_1 \ldots C_p$ be the children of $N$ in topologically sorted order;
    $\mathbf{x}_0 := G[N]$;
    $q := N.\text{label}$;
    pick points $x_1 \ldots x_p$ in increasing order such that
        for all $i, j \in 0 \ldots p$, if $i \neq j$ then $\mathrm{od}(x_i, x_j) = \delta_q$;
        /* Such points can be chosen by virtue of axiom A.8 */
    $\mathbf{for}$ $i = 1 \ldots p$ do
        $G[C_i] := x_i$;
        instantiate1($C_i, \Omega, \delta_1 \ldots \delta_k, G$)
    $\mathbf{endfor}$
$\mathbf{endif\ end}$ instantiate1.

Algorithm solve_constraints is modified as follows:

$\mathbf{procedure}$ solve_constraints2($\mathbf{in}$ $\mathcal{S}$: a system of constraints of the form $\mathrm{od}(a, b) \ll \mathrm{od}(c, d)$ ;
{NEW}                      $\mathcal{O}$ : a system of constraints of the form $a < b$)
                  $\mathbf{return\ either}$ an ordered cluster tree $T$ satisfying $\mathcal{S}$
                            if $\mathcal{S}$ is consistent;
                      $\mathbf{or\ false}$ if $\mathcal{S}$ is inconsistent.

$\mathbf{variables:}$ $m$ is an integer;
        $C$ is a constraint in $\mathcal{S}$;
        $H, I$ are undirected graphs;
        $M, N, P$ are nodes of $T$;
        $a, b, c, d$ are symbols;

$\mathbf{begin}$  $\mathbf{if}$ $\mathcal{S}$ contains any constraint of the form, "$\mathrm{od}(a, b) \ll \mathrm{od}(c, c)$"
        $\mathbf{then\ return\ false}$;
{NEW}$\mathbf{if}$ $\mathcal{O}$ is internally inconsistent (contains a cycle) $\mathbf{then\ return\ false}$;
        $m :=$ the number of variables in $\mathcal{S}$;
        initialize $T$ to consist of a single node $N$;
        $N.\text{symbols} :=$ the variables in $\mathcal{S}$;

        $\mathbf{repeat}$ $H :=$ the connected component of the shorts of $\mathcal{S}$;

{NEW}    $H$ := incorporate_order$(H, \mathcal{O})$;
    **if** $H$ contains all the edges in $\mathcal{S}$ **then return false**

    **for** each leaf $N$ of $T$ do
      **if not** all vertices of $N$ are connected in $H$ **then**
        $N$.label := $m$;
        **for** each connected component $I$ of $N$.symbols in $H$ **do**
          construct node $M$ as a new child of $N$ in $T$;
          $M$.symbols:= the vertices of $I$;
        **endfor endif**
{NEW}      **for** each constraint $a < b \in \mathcal{O}$
{NEW}        **if** $a$ is in $M$.symbols and $b$ is in $P$.symbols
{NEW}          where $M$ and $P$ are different children of $N$
{NEW}          **then** add an ordering arc from $M$ to $P$;
{NEW}          **endif endfor**
    **endfor**

    $\mathcal{S}$ := the subset of constraints in $\mathcal{S}$ whose long is in $H$;
    $m := m - 1$;
    **until** $\mathcal{S}$ is empty;

  **for** each leaf $N$ of $T$
    $N$.label := 0;
    **if** $N$.symbols has more than one symbol
      **then** create a leaf of $N$ for each symbol in $N$.symbols;
        label each such leaf 0;
      **endif endfor**
**end** solve_constraints2.

{NEW}
**function** incorporate_order(**in** $H$ : undirected graph;
         $\mathcal{O}$ : a system of constraints of the form $a < b$)
      **return** undirected graph;

**variables:** $G$ : directed graph;
    $a, b$ : vertices in $H$;
    $A, B$ : connected components of $H$;
    $V[A]$ : **array** of vertices of $G$ indexed on connected components of $H$;
    $I$ : subset of vertices of $G$;

**for each** connected component $A$ of $H$ create a vertex $V[A]$ in $G$;
**for each** constraint $a < b \in \mathcal{O}$
    let $A$ and $B$ be the connected components of $H$ containing $a$ and $b$ respectively;
    **if** $A \neq B$ **then** add an arc in $G$ from $V[A]$ to $V[B]$ **endif endfor**;
**for each** strongly connected component $I$ of $G$ **do**
    **for each** pair of distinct vertices $V[A], V[B] \in I$ **do**
      **for each** $a \in A$ and $b \in B$ add the edge $ab$ to $H$ **endfor endfor**
**endfor**

**end** incorporate_order.

Function incorporate_order serves the following purpose. Suppose that we are in the midst of the main loop of solve_constraints2, we have a partially constructed cluster tree, and we are currently working on finding the sub-clusters of a node $N$. As in the original form of solve_constraints, we find the connected components of the shorts of the order-of-magnitude constraints. Let these be $C_1 \ldots C_q$; then we know that the diameter of each $C_i$ is much smaller than the diameter of $N$. Now, suppose, for example, that we have in $\mathcal{O}$ the constraints $a_1 < a_5, b_5 < b_2, c_2 < c_1$, where $a_1, c_1 \in C_1$; $b_2, c_2 \in C_2$; and $a_5, b_5 \in C_5$. Then it follows from axiom A.9 that $C_1$, $C_2$, and $C_5$ must all be merged into a single cluster, whose diameter will be less than the diameter of $N$. Procedure incorporate_order finds all such loops by constructing a graph $G$ whose vertices are the connected components of $H$ and whose arcs are the ordering relations in $\mathcal{O}$ and then computing the strongly connected components of $G$. (Recall that two vertices $u, v$ in a directed graph are in the same strongly connected component if there is a cycle from $u$ to $v$ to $u$.) It then merges together all of the connected components of $H$ that lie in a single strongly connected component of $G$.

The proof of the correctness of algorithm solve_constraints2 is again analogous in structure to the proof of theorem 1, and is given in the appendix.

By implementing this in the manner of Section 6.1, the algorithm can be made to run in time $O(\max(n^2\alpha(n), ne, no, s))$, where $o$ is the number of constraints in $\mathcal{O}$.

# 7. Finite order of magnitude comparison

In this section, it is demonstrated that algorithm solve_constraints can be applied to systems of constraints of the form "dist$(a, b) <$ dist$(c, d)$ $/$ $B$" for finite $B$ in ordinary Euclidean space as long as the number of symbols in the constraint network is smaller than $B$.

We could be sure immediately that some such result must apply for finite $B$. It is a fundamental property of the non-standard real line that any sentence in the first-order theory of the reals that holds for all infinite values holds for any sufficiently large finite value, and that any sentence that holds for some infinite value holds for arbitrarily large finite values. Hence, since the answer given by algorithm solve_constraints works over a set of constraints $\mathcal{S}$ when the constraint "od$(a, b) \ll$ od$(c, d)$" is interpreted as "od$(a, b) <$ od$(c, d)/B$ for infinite $B$", the same answer must be valid for sufficiently large finite $B$. What is interesting is that we can find a simple characterization of $B$ in terms of $\mathcal{S}$; namely, that $B$ is larger than the number of symbols in $\mathcal{S}$.

We begin by modifying the form of the constraints, and the interpretation of a cluster tree. First, to avoid confusion, we will use a four-place predicate "much_closer$(a, b, c, d)$" rather than the form "od$(a, b) \ll$ od$(c, d)$" as we are not going to give an interpretation to "od" as a function. We fix a finite value $B > 1$, and interpret "much_closer$(a, b, c, d)$" to mean "dist$(a, b) <$ dist$(c, d)$ $/$ $B$."

We next redefine what it means for a valuation to instantiate a cluster tree:

**Definition 6:** Let $T$ be a cluster tree and let $\Gamma$ be a valuation on the symbols in $T$. We say that $\Gamma \vdash T$ if the following holds: For any symbols $a, b, c, d$ in $T$, let $M$ be the least common ancestor of $a, b$ and let $N$ be the least common ancestor of $c, d$. If $M$.label $< N$.label then much_closer$(a, b, c, d)$.

Procedure "instantiate", which generates an instantiation of a cluster tree, is modified as follows:

**procedure** instantiate(**in** $T$ : cluster tree; $\Omega$ : Euclidean space; $B$ : real);
                **return** : array of points indexed on the symbols of $T$;

Let $n$ be the number of nodes in $T$;
$\alpha := 2 + 2n + Bn$;
Choose $\delta_1, \delta_2 \ldots \delta_n$ such that $\delta_i < \delta_{i+1}/\alpha$;
pick a point $x \in \Omega$;
$G[T] := x$;
instantiate1$(T, \Omega, \delta_1 \ldots \delta_n, G)$;
**return** the restriction of $G$ to the symbols of $T$.
**end** instantiate.

instantiate1(**in** $N$ : a node in a cluster tree; $\Omega$ : a Euclidean space;
                $\delta_1 \ldots \delta_n$ : orders of magnitude;
              **in out** G : array of points indexed on the nodes of $T$)
**if** $N$ is not a leaf **then**
    let $C_1 \ldots C_p$ be the children of $N$;
    $\mathbf{x}_1 := G[N]$;
    $q := N$.label;
    pick points $x_2 \ldots x_p$ such that
        for all $i, j \in 1 \ldots p$, if $i \neq j$ then $\delta_q \leq \mathrm{dist}(x_i, x_j) < n\delta_q$
    /* This is possible since $p \leq n$. */
    **for** $i = 1 \ldots p$ **do**
        $G[C_i] := x_i$;
        instantiate1$(C_i, \Omega, \delta_1 \ldots \delta_n, G)$
    **endfor**
**endif end** instantiate1.

The analogue of lemma 2 holds for the revised algorithm:
**Lemma 22:** Any cluster tree $T$ has an instantiation in Euclidean space $\Re^m$ of any dimensionality $m$.

We can now state theorem 3, which asserts the correctness of algorithm "solve_constraints" in this new setting:

**Theorem 3:** Let $\mathcal{S}$ be a set of constraints over $n$ variables of the form "dist$(a, b)$ < dist$(c, d)$ / $B$", where $B > n$. The algorithm solve_constraints($\mathcal{S}$) returns a cluster tree satisfying $\mathcal{S}$ if $\mathcal{S}$ is consistent over Euclidean space, and returns **false** if $\mathcal{S}$ is inconsistent.

The proofs of lemma 22 and theorem 3 are given in the appendix.

An examination of the proof of lemma 22 shows that this result does not depend on any relation between $n$ and $B$. Therefore, if solve_constraints($\mathcal{S}$) returns a tree $T$, then $\mathcal{S}$ is consistent and $T$ satisfies $\mathcal{S}$ regardless of the relation between $n$ and $B$. However, it is possible for $\mathcal{S}$ to be consistent and solve_constraints($\mathcal{S}$) to return **false** if $n \geq B$. On the other hand, one can see from the proof of theorem 3 (particularly lemma 23) that if $B > n$ and solve_constraints($\mathcal{S}$) returns **false** then $\mathcal{S}$ is inconsistent in any metric space. However, there are metric spaces other than $\Re^m$ in which the cluster tree returned by solve_constraints may have no instantiation.

## 8. The first-order theory

Our final result asserts that if the om-space is rich enough then the full first-order language of order-of-magnitude distance comparisons is decidable. Specifically, if the collection of orders of magnitude is dense and unbounded above, then there is a decision algorithm for first-order sentences over the formula, "od$(W, X) \ll$ od$(Y, Z)$" that runs in time $O(4^n (n!)^2 s)$ where $n$ is the number of variables in the sentence and $s$ is the length of the sentence.

The basic reason for this is the following: As we have observed in corollary 4, a cluster tree $T$ determines the truth value of all constraints of the form "od$(a, b) \ll$ od$(c, d)$" where $a, b, c, d$ are symbols in the tree. That is, any two instantiations of $T$ in any two om-spaces agree on any such constraint. If we further require that the om-spaces are dense and unbounded, then a much stronger statement holds: Any two instantiations of $T$ over such om-spaces agree on *any* first-order formula free in the symbols of $T$ over the relation "od$(W, X) \ll$ od$(Y, Z)$". Hence, it suffices to check the truth of a sentence over all possible cluster trees on the variables in the sentence. Since there are only finitely many cluster trees over a fixed set of variables (taking into account only the relative order of the labels and not their numeric values), this is a decidable procedure.

Let $\mathcal{L}$ be the first-order language with equality with no constant or function symbols, and the single predicate symbol "much_closer$(a, b, c, d)$". It is easily shown that $\mathcal{L}$ is as expressive as the language with the function symbol "od" and the relation symbol $\ll$.

**Definition 7:** An om-space $\Omega$ with orders of magnitude $\mathcal{D}$ is *dense* if it satisfies the following axiom:

A.9 For all orders of magnitude $\delta_1 \ll \delta_3$ in $\mathcal{D}$, there exists a order of magnitude $\delta_2$ in $\mathcal{D}$ such that $\delta_1 \ll \delta_2 \ll \delta_3$.

$\Omega$ is *unbounded above* if it satisfies the following:

A.10 For every order of magnitude $\delta_1$ in $\mathcal{D}$ there exists $\delta_2$ in $\mathcal{D}$ such that $\delta_1 \ll \delta_2$.

If $\mathcal{D}$ is the collection of orders of magnitude in the hyperreal line, then both of these are satisfied. In axiom [A.9], if $0 \ll \delta_1 \ll \delta_3$, choose $\delta_2 = \sqrt{\delta_1 \delta_3}$, the geometric mean. If $0 = \delta_1 \ll \delta_3$, choose $\delta_2 = \delta_3 \delta$ where $\delta \ll 1$. In axiom [A.10] choose $\delta_2 = \delta_1 / \delta$ where $0 < \delta \ll 1$.

**Definition 8:** Let $T$ be a cluster tree. Let $l_0 = 0, l_1, l_2 \ldots l_k$ be the distinct labels in $T$ in ascending order. An *extending label* for $T$ is either (a) $l_i$ for some $i$; (b) $l_k + 1$ (note that $l_k$ is the label of the root); (c) $(l_{i-1} + l_i)/2$ for some $i$ between 1 and $k$.

Note that if $T$ has $k$ distinct non-zero labels, then there are $2k + 2$ different extending labels for $T$.

**Definition 9:** Let $T$ be a cluster tree. Let $x$ be a symbol not in $T$. The cluster tree $T'$ *extends $T$ with $x$* if $T'$ is formed from $T$ by applying one of the following operations (a single application of a single operation).

1. $T$ is the null tree and $T'$ is the tree containing the single node $x$.

2. $T$ consists of the single node for symbol $y$. Make a new node $M$, make both $x$ and $y$ children of $M$, and set the label of $M$ to be either 0 or 1.

3. For any internal node $N$ of $T$ (including the root), make $x$ a child of $N$.

4. Let $y$ be a symbol in $T$, and let $N$ be its father. If $N$.label $\neq 0$, create a new node $M$ with an extending label for $T$ such that $M$.label $<$ $N$.label. Make $M$ a child of $N$, and make $x$ and $y$ children of $M$.

5. Let $C$ be an internal node of $T$ other than the root, and let $N$ be its father. Create a new node $M$ with an extending label for $T$ such that $C$.label $<$ $M$.label $<$ $N$.label. Make $M$ a child of $N$ and make $x$ and $C$ children of $M$.

6. Let $R$ be the root of $T$. Create a new node $M$ such that $M$.label $= R$.label $+ 1$. Make $R$ and $x$ children of $M$. Thus $M$ is the root of the new tree $T'$.

(See Figure 3.)

Note that if $T$ is a tree of $n$ symbols and at most $n - 1$ internal nodes then

- There are $n - 1$ ways to carry out step 3.

- There are $n$ possible ways to choose symbol $y$ in step 4, and at most $2n - 2$ for the label on $M$ in each.

- There are at most $n - 2$ different choices for $C$ in step 5, and at most $2n - 3$ choices for the label on $M$ in each.

- There is only one way to carry out step 6.

Hence, there are less than $4n^2$ different extensions of $T$ by $x$. (This is almost certainly an overestimate by at least a factor of 2, but the final algorithm is so entirely impractical that it is not worthwhile being more precise.)

**Definition 10:** Let $T$ be a cluster tree, and let $\phi$ be a formula of $\mathcal{L}$ open in the variables of $T$. $T$ *satisfies* $\phi$ if every instantiation of $T$ satisfies $\phi$.

**Theorem 4:** Let $T$ be a cluster tree. Let $\phi$ be an open formula in $\mathcal{L}$, whose free variables are the symbols of $T$. Let $\Omega$ be an om-space that is dense and unbounded above. Algorithm decide$(T, \phi)$ returns **true** if $T$ satisfies $\phi$ and **false** otherwise.

**function** decide$(T$ : cluster tree; $\phi$ : formula$)$ **return boolean**
convert $\phi$ to an equivalent form in which the only logical symbols in $\phi$ are
$\qquad \neg$ (not), $\wedge$ (and), $\exists$ (exists), $=$ (equals) and variable names,
$\qquad$ and the only non-logical symbol is the predicate "much_closer".
**case**
$\qquad \phi$ has form $X = Y$: **return** (distance$(X, Y, T) = 0$);
$\qquad \phi$ has form "much_closer$(W, X, Y, Z)$": **return** distance$(W, X, T) <$ distance$(Y, Z, T)$);
$\qquad \phi$ has form $\neg \psi$: **return not**(decide$(T, \psi)$)
$\qquad \phi$ has form $\psi \wedge \theta$: **return**(decide$(T, \psi)$ **and** decide$(T, \theta)$)
$\qquad \phi$ has form $\exists_X \alpha$;

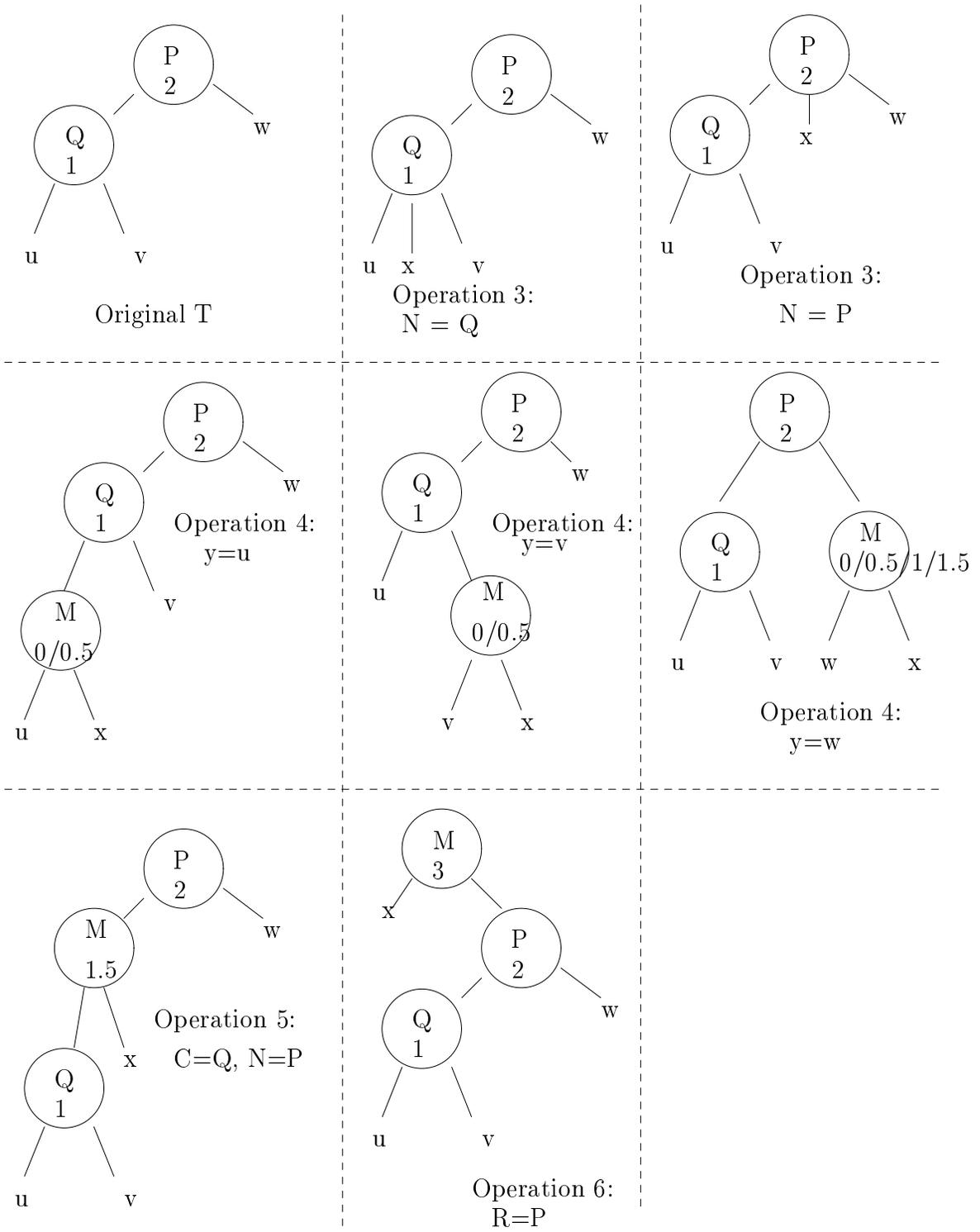

Figure 3: Extensions of a cluster tree

```
        if for some extension T' of T by X, decide(T', α) = true
            then return true
            else return false endif endcase
end decide
```

```
function distance(X, Y : symbol; T : cluster tree) return integer
N := the common ancestor of X and Y in T;
return(N.label)
end distance
```

The proof of theorem 4 is given in the appendix.

Running time: As we have remarked above, for a tree $T$ of size $k$ there are at most $4k^2$ extensions of $T$ to be considered. The total number of cluster trees considered is therefore bounded by $\Pi_{k=1}^{n} 4k^2 = 4^n (n!)^2$. It is easily verified that the logical operators other than quantifiers add at most a factor of $s$ where $s$ is the length of the sentence. Hence the running time is bounded by $O(4^n (n!)^2 s)$.

A key lemma, of interest in itself, states the following:

**Lemma 28:** Let $T$ be a cluster tree. Let $\phi$ be an open formula in $\mathcal{L}$, whose free variables are the symbols of $T$. Let $\Omega$ be an om-space that is dense and unbounded above. If one instantiation $\Gamma$ of $T$ in $\Omega$ satisfies $\phi$ then every instantiation of $T$ in $\Omega$ satisfies $\phi$.

That is, either $\phi$ is true for all instantiations of $T$ or for none. The proof is given in the appendix.

It should be observed that the above conditions on $\Omega$ in lemma 28 are necessary, and that the statement is false otherwise. For example, let $\Omega$ be the om-space described in example I, Section 3, of polynomials over an infinitesimal $\delta$. Then $\Omega$ is not unbounded above; there is a maximum order-of-magnitude $O(1)$. Let $T$ be the starting tree of Figure 3 (upper-left corner). Let $\phi$ be the formula "$\exists_X \text{ od}(V, W) \ll \text{od}(W, X)$", free in $V$ and $W$. Then the valuation $\{U \to \delta, V \to 0, W \to 1\}$ satisfies $T$ but not $\phi$, whereas the valuation $\{U \to \delta^2, V \to 2\delta^2, W \to \delta\}$ satisfies both $T$ and $\phi$.

## 9. Conclusions

The applications of the specific algorithms above are undoubtedly limited; we are not aware of any practical problems where solving systems of order-of-magnitude relations on distances is the central problem. However, the potential applications of order-of-magnitude reasoning generally are very widespread. Ordinary commonsense reasoning involves distances spanning a ratio of about $10^8$, from a fraction of an inch to thousands of miles, and durations spanning a ratio of about $10^{10}$, from a fraction of a second to a human lifetime. Scientific reasoning spans much greater ranges. Explaining the dynamics of a star combines reasoning about nuclear reactions with reasoning about the star as a whole; these differ by a ratio of about $10^{57}$. The techniques needed to compute with quantities of such vastly differing sizes are quite different from the techniques needed to compute with quantities all of similar sizes. This paper is a small step in the development and analysis of such computational techniques.

The above results are also significant in the encouragement that they give to the hope that order-of-magnitude reasoning specifically, and qualitative reasoning generally, may lead to useful quick reasoning strategies in a broader range of problems. It has been often found in AI that moving from greater to lesser precision in the mode of inference or type of knowledge does not lead to quick and dirty heuristic techniques, but rather to slow and dirty techniques. Nonmonotonic reasoning is the most notorious example of this, but it arises as well in many other types of automated reasoning, including qualitative spatial and physical reasoning. The algorithms developed in this paper are a welcome exception to this rule. We are currently studying algorithmic techniques for other order-of-magnitude problems, and are optimistic of finding similar favorable results.

## Acknowledgements

This research has been supported by NSF grant #IRI-9625859. Thanks to Ji-Ae Shin, Andrew Gelsey, and the reviewers for helpful comments.

## Appendix A. Proofs

In this appendix, we give the proofs of the various results asserted in the body of the paper.

### Proof of Lemma 2

**Lemma 2:** If $T$ is a cluster tree and $\Omega$ is an om-space, then instantiate($T, \Omega$) returns an instantiation of $T$.

**Proof:** Let $\delta_0 = 0$. For any node $N$, if $i = N$.label, we define $\Delta(N) = \delta_i$. The proof then proceeds in the following steps:

i. For any nodes $M, N$, if $M$ is a descendant of $N$ in $T$ then od($G[M], G[N]$) $\lll \Delta(N)$.
   **Proof:** If $M$ is a child of $N$, then this is immediate from the construction of $x_2 \dots x_p$ in instantiate1. Else, let $N = N_1, N_2 \dots N_q = M$ be the path from $N$ to $M$ through $T$. By the definition of a cluster tree, it follows that $N_i$.label $< N$.label, for $i > 1$ and therefore $\Delta(N_i) \ll \Delta(N)$. Thus od($G[M], G[N]$) $\lll$ (by the o.m.-triangle inequality) $\max_{i=1\dots q-1}(\text{od}(G[N_{i+1}], G[N_i])) \lll \max_{i=1\dots q-1}(\Delta(N_i))$ (since $N_{i+1}$ is the child of $N_i$) $\lll \Delta(N)$.

ii. Let $N$ be a node in $T$; let $C_1$ and $C_2$ be two distinct children of $N$; and let $M_1$ and $M_2$ be descendants of $C_1$ and $C_2$ respectively. Then od($G[M_1], G[M_2]$) $= \Delta(N)$.
   **Proof:** By the construction of $x_2 \dots x_p$ in instantiate1($N$), od($G[C_1], G[C_2]$) $= \Delta(N)$. By part (i.), od($G[M_1], G[C_1]$) $\lll \Delta(C_1) \ll \Delta(N)$ and likewise od($G[M_2], G[C_2]$) $\ll \Delta(N)$. Hence, by axiom A.6, od($G[M_1], G[M_2]$) $= \Delta(N)$.

iii. Let $a$ and $b$ be any two leaves in $T$, and let $N$ be the least common ancestor in $T$ of $a$ and $b$. Then od($G[a], G[b]$) $= \Delta(N)$. **Proof:** Immediate from (ii).

iv. For any node $N$, odiam($\Gamma(N)$) $= \Delta(N)$. **Proof:** From (iii), any two leaves descending from different children of $N$ are at a distance of order $\Delta(N)$, and no two leaves of $N$ are at a distance of order greater than $\Delta(N)$.

v. For any node $N$, $\Gamma(N)$ is a cluster of $\Gamma(T)$. **Proof:** Let $a$ and $b$ be leaves of $N$, and let $c$ be a leaf of $T - N$. Let $I$ be the common ancestor of $a$ and $b$ in $T$ and let $J$ be the common ancestor of $a$ and $c$. Then $I$ is either $N$ or a descendant of $N$ and $J$ is a proper ancestor of $N$. Therefore by part (i), $\Delta(I) \ll \Delta(J)$. But by (iii), $\text{od}(\Gamma(a), \Gamma(b)) = \Delta(I) \ll \Delta(J) = \text{od}(\Gamma(a), \Gamma(c))$.

vi. For any internal nodes $N, M$ if $M.\text{label} < N.\text{label}$ then $\text{odiam}(\Gamma(M)) \ll \text{odiam}(\Gamma(N))$. **Proof:** Immediate from (iv) and the construction of $\Delta$.

vii. If $C$ is a cluster of $\Gamma(T)$ then there is a node $N$ in $T$ such that $C = \Gamma(N)$. **Proof:** Let $S$ be the set of symbols corresponding to $C$ and let $N$ be the least common ancestor of all of $S$. Let $a$ and $b$ be two symbols in $S$ that are in different subtrees of $N$. Then by (iii), $\text{od}(G[a], G[b]) = \Delta(N)$. Let $x$ be any symbol in $N.\text{symbols}$. Then by (iii) $\text{od}(G[a], G[x]) \lleq \Delta(N)$. Hence $G[x] \in C$.

□

## Proof of Theorem 1

We here prove the correctness of algorithm solve_constraints. We will assume throughout that the two variables in the long of any constraint in $\mathcal{S}$ are distinct.

**Lemma 3:** Let $T$ be a cluster tree and let $\Gamma$ be an instantiation of $T$. Let $a$ and $b$ be symbols of $T$. Let $N$ be the least common ancestor of $a$ and $b$ in $T$. Then $\text{od}(\Gamma(a), \Gamma(b)) = \text{odiam}(\Gamma(N))$.

**Proof:** Since $\Gamma(a)$ and $\Gamma(b)$ are elements of $\Gamma(N)$, it follows from the definition of odiam that $\text{od}(\Gamma(a), \Gamma(b)) \lleq \text{odiam}(\Gamma(N))$. Suppose the inequality were strict; that is, $\text{od}(\Gamma(a), \Gamma(b)) \ll \text{odiam}(\Gamma(N))$. Then let $C$ be the set of all the symbols $c$ of $T$ such that $\text{od}(\Gamma(a), \Gamma(c)) \lleq \text{od}(\Gamma(a), \Gamma(b))$. Then $\text{odiam}(\Gamma(C)) = \text{od}(\Gamma(a), \Gamma(b)) \ll \text{odiam}(\Gamma(N))$. It is easily shown that $\Gamma(C)$ is a cluster in $\Gamma(T)$. Therefore, by property (ii) of definition 5, there must be a node $M$ such that $M.\text{symbols} = C$. Now, $M$ is certainly not an ancestor of $N$, since $\text{odiam}(\Gamma(M)) \ll \text{odiam}(\Gamma(N))$ but $M.\text{symbols}$ contains both $a$ and $b$. But this contradicts the assumption that $N$ was the least common ancestor of $a$ and $b$. □

**Corollary 4:** Let $T$ be a cluster tree and let $\Gamma$ be an instantiation of $T$. Let $a, b, c, d$ be symbols of $T$. Let $N$ be the least common ancestor of $c$ and $d$ in $T$, and let $M$ be the least common ancestor of $a$ and $b$ in $T$. Then $\text{od}(\Gamma(a), \Gamma(b)) \ll \text{od}(\Gamma(c), \Gamma(d))$ if and only if $M.\text{label} < N.\text{label}$.

**Proof:** Immediate from lemma 3 and property (iii) of definition 5 of instantiation. □

**Lemma 5:** Let $\mathcal{S}$ be any set of constraints of the form $\text{od}(a, b) \ll \text{od}(c, d)$. Let $H$ be the connected components of the shorts of $\mathcal{S}$. If $\mathcal{S}$ is consistent, then not every edge of $\mathcal{S}$ is in $H$.

**Proof:** Let $\Gamma$ be a valuation satisfying $\mathcal{S}$. Find an edge $pq$ in $\mathcal{S}$ for which $\text{od}(\Gamma(p), \Gamma(q))$ is maximal. Now, if $ab$ is a short of $\mathcal{S}$ — that is, there is a constraint $\text{od}(a, b) \ll \text{od}(c, d)$ in $\mathcal{S}$ — then $\text{od}(\Gamma(a), \Gamma(b)) \ll \text{od}(\Gamma(c), \Gamma(d)) \lleq \text{od}(\Gamma(p), \Gamma(q))$.

Now, let $ab$ be any edge in $H$, the connected components of the shorts of $\mathcal{S}$. Then there is a path $a_1 = a, a_2 \ldots a_k = b$ such that the edge $a_i a_{i+1}$ is a short of $\mathcal{S}$ for $i = 1 \ldots k - 1$. Thus, by the om-triangle inequality, $\mathrm{od}(\Gamma(a), \Gamma(b)) \lessapprox \max_{i=1..k-1}(\mathrm{od}(\Gamma(a_i), \Gamma(a_{i+1}))) \ll \mathrm{od}(\Gamma(p), \Gamma(q))$. Hence $pq \neq ab$, so $pq$ is not in $H$. $\square$

**Lemma 6:** The values of $\mathcal{S}$ and $H$ in any iteration are supersets of their values in any later iteration.

**Proof:** $\mathcal{S}$ is reset to a subset of itself at the end of each iteration. $H$ is defined in terms of $\mathcal{S}$ in a monotonic manner. $\square$

**Lemma 7:** $\mathcal{S}$ cannot be the same in two successive iterations of the main loop.

**Proof:** by contradiction. Suppose that $\mathcal{S}$ is the same in two successive iterations. Then $H$ will be the same, since it is defined in terms of $\mathcal{S}$. $H$ is constructed to contain all the shorts of $\mathcal{S}$, Since the resetting of $\mathcal{S}$ at the end of the first iteration does not change $\mathcal{S}$, $H$ must contain all the longs as well. Thus, $H$ contains all the edges in $\mathcal{S}$. But that being the case, the algorithm should have terminated with failure at the beginning of the first iteration. $\square$

**Lemma 8:** Algorithm solve_constraints always terminates.

**Proof:** By lemma 7, if the algorithm does not exit with failure, then on each iteration some constraints are removed from $\mathcal{S}$. Hence, the number of iterations of the main loop is at most the original size of $\mathcal{S}$. Everything else in the algorithm is clearly bounded. (Note that this bound on the number of iterations is improved in Section 6.1 to $n - 1$, where $n$ is the number of symbols.) $\square$

**Lemma 9:** If algorithm solve_constraints returns **false**, then $\mathcal{S}$ is inconsistent.

**Proof:** If the algorithm returns **false**, then the transitive closure of the shorts of $\mathcal{S}$ contains all the edges in $\mathcal{S}$. By lemma 5, $\mathcal{S}$ is inconsistent.

**Lemma 10:** If constraint $C$ of form $\mathrm{od}(a, b) \ll \mathrm{od}(c, d)$ is in the initial value of $\mathcal{S}$, and edge $cd$ is in $H$ in some particular iteration, then constraint $C$ is in $\mathcal{S}$ at the start of that iteration.

**Proof:** Suppose that $C$ is deleted from $S$ on some particular iteration. Then edge $cd$, the long of $C$, cannot be in $H$ in that iteration. That is, it is not possible for edge $cd$ to persist in $H$ in an iteration after $C$ has been deleted from $\mathcal{S}$. Note that, by lemma 6, once $cd$ is eliminated from $H$, it remains out of $H$. $\square$

**Lemma 11:** The following loop invariant holds: At the end of each loop iteration, the values of $L$.symbols, where $L$ is a leaf in the current state of the tree, are exactly the connected components of $H$.

**Proof:** In the first iteration, $T$ is initially just the root $R$, containing all the symbols, and a child of $R$ is created for each connected component of $H$.

Let $T_i$ and $H_i$ be the values of $T$ and $H$ at the end of the $i$th iteration. Suppose that the invariant holds at the end of the $k$th iteration. By lemma 6, $H_{k+1}$ is a subset of $H_k$. Hence, each connected component of $H_{k+1}$ is a subset of a connected component of $H_k$.

Moreover, each connected component $J$ of $H_k$ is either a connected component of $H_{k+1}$ or is partitioned into several connected components of $H_{k+1}$. In the former case, the leaf of $T_k$ corresponding to $J$ is unchanged and remains a leaf in $T_{k+1}$. In the latter case, the leaf corresponding to $J$ gets assigned one child for each connected component of $H_{k+1}$ that is a subset of $J$. Thus, the connected components of $H_{k+1}$ correspond to the leaves of $T_{k+1}$. $\square$

**Lemma 12:** If procedure solve_constraints does not return **false**, then it returns a well-formed cluster tree $T$.

**Proof:** Using lemma 11, and the cleanup section of solve_constraints which creates the final leaves for symbols, it follows that every symbol in $\mathcal{S}$ ends up in a single leaf of $T$. As $m$ is decremented on each iteration, and as no iteration adds both a new node and children of that node, it follows that the label of each internal node is less than the label of its father. Hence the constraints on cluster trees (definition 3) are satisfied. $\square$

**Lemma 13:** Let $a, b$ be two distinct symbols in $\mathcal{S}$ and let $T$ be the cluster tree returned by solve_constraints for $\mathcal{S}$. Let $N$ be the least common ancestor of $a, b$ in $T$. Then either $N$ is assigned its label on the first iteration when the edge $ab$ is not in $H$, or the edge $ab$ is in the final value of $H$ when the loop is exited and $N$ is assigned its label in the final cleanup section.

**Proof:** As above, let $H_i$ be the value of $H$ in the $i$th iteration.

If $N$ is the root, then it is assigned its label in the first iteration. Clearly, $a$ and $b$, being in different subtrees of $N$, must be in different connected components of $H_1$.

Suppose $N$ is assigned its label in the $k$th iteration of the loop for $k > 1$. By lemma 11, at the end of the previous iteration, $N$.symbols was a connected component of $H_{k-1}$, and it therefore contained the edge $ab$. Since $N$ is the least common ancestor of $a, b$, it follows that $a$ and $b$ are placed in two different children of $N$; hence, they are in two different connected components of $H_k$. Thus the edge $ab$ cannot be in $H_k$.

Suppose $N$ is assigned its label in the cleanup section of the algorithm. Then by lemma 11, $N$.symbols is a connected component of the final value of $H$. Hence the edge $ab$ was in the final value of $H$. $\square$

**Lemma 14:** Let $\mathcal{S}$ initially contain constraint $C$ of form $\text{od}(a, b) \ll \text{od}(c, d)$. Suppose that solve_constraints($\mathcal{S}$) returns a cluster tree $T$. Let $M$ be the least common ancestor of $a, b$ in $T$ and let $N$ be the least common ancestor of $c, d$. Then $M$.label $< N$.label.

**Proof:** Suppose $N$ is given a label in a given iteration. By lemma 13, $cd$ is eliminated from $H$ in that same iteration. By lemma 10, constraint $C$ must be in $\mathcal{S}$ at the start of the iteration. Hence $ab$ is a short of $S$ in the iteration, and is therefore in $H$. Hence $M$ is not given a label until a later iteration, and therefore is given a lower label.

It is easily seen that $cd$ cannot be in $H$ in the final iteration of the loop, and hence $N$ is not assigned its label in the cleanup section. $\square$

**Lemma 15:** Suppose that solve_constraints($\mathcal{S}$) returns a cluster tree $T$. Then any instantiation of $T$ satisfies the constraints $\mathcal{S}$.

**Proof:** Immediate from lemma 14 and corollary 4.

**Theorem 1:** The algorithm solve_constraints($\mathcal{S}$) returns a cluster tree satisfying $\mathcal{S}$ if $\mathcal{S}$ is consistent, and returns **false** if $\mathcal{S}$ is inconsistent.

**Proof:** If solve_constraints($\mathcal{S}$) returns **false**, then it is inconsistent (lemma 9). If it does not return **false**, then it returns a cluster tree $T$ (lemma 12). Since $T$ has an instantiation (lemma 2) and since every instantiation of $T$ is a solution of $\mathcal{S}$ (lemma 15), it follows that $\mathcal{S}$ is consistent and $T$ satisfies $\mathcal{S}$. $\square$

**Proof of Theorem 2**

**Lemma 16:** If $\mathcal{S}_1$ and $\mathcal{S}_2$ are consistent sets of constraints, and $\mathcal{S}_1 \supset \mathcal{S}_2$ then reduce_constraints($\mathcal{S}_1$) $\supseteq$ reduce_constraints($\mathcal{S}_2$).

**Proof:** Immediate by construction. The value of $H$ in the case of $\mathcal{S}_1$ is a superset of its value in the case of $\mathcal{S}_2$, and hence reduce_constraints($\mathcal{S}_1$) is a superset of reduce_constraints($\mathcal{S}_2$).

**Lemma 17:** If $\mathcal{S}_1$ and $\mathcal{S}_2$ are consistent sets of constraints, and $\mathcal{S}_1 \supset \mathcal{S}_2$ then num_labels($\mathcal{S}_1$) $\geq$ num_labels($\mathcal{S}_2$).

**Proof** by induction on num_labels($\mathcal{S}_2$). If num_labels($\mathcal{S}_2$) $= 0$, the statement is trivial. Suppose that the statement holds for all $\mathcal{S}'$, where num_labels($\mathcal{S}'$) $= k$.
Let num_labels($\mathcal{S}_2$) $= k + 1$.
Then $k + 1 =$ num_labels($\mathcal{S}_2$) $= 1 +$ num_labels(reduce_constraints($\mathcal{S}_2$)), so $k =$num_labels(reduce_constraints($\mathcal{S}_2$)). Now, suppose $\mathcal{S}_1 \supset \mathcal{S}_2$. By lemma 16 reduce_constraints($\mathcal{S}_1$) $\supset$ reduce_constraints($\mathcal{S}_2$). But then by the inductive hypothesis num_labels(reduce_constraints($\mathcal{S}_1$)) $\geq$ num_labels(reduce_constraints($\mathcal{S}_2$)), so num_labels($\mathcal{S}_1$) $\geq$ num_labels($\mathcal{S}_2$). $\square$

**Lemma 18:** Let $\mathcal{S}$ be a set of constraints, and let $\Gamma$ be a solution of $\mathcal{S}$. For any graph $G$ over the symbols of $\mathcal{S}$, let nd($G, \Gamma$) be the number of different non-zero values of od($a, b$) where edge $ab$ is in $G$. Let edges($\mathcal{S}$) be the set of edges in $\mathcal{S}$. Then nd(edges($\mathcal{S}$), $\Gamma$) $\geq$ num_labels($\mathcal{S}$).

**Proof:** by induction on num_labels($\mathcal{S}$). If num_labels($\mathcal{S}$) $= 0$, then the statement is trivial. Suppose for some $k$, the statement holds for all $\mathcal{S}'$ where num_labels($\mathcal{S}'$) $= k$, and suppose num_labels($\mathcal{S}$) $= k + 1$. Let $pq$ be the edge in $\mathcal{S}$ of maximal length. For any set of edges $E$, let small-edges($E, \Gamma$) be the set of all edges $ab$ in $E$ for which od($\Gamma(a), \Gamma(b)$) $\ll$ od($\Gamma(p), \Gamma(q)$). Since small-edges($E$) contains edges of every order of magnitude in $E$ except the order of magnitude of $pq$, it follows that nd(small-edges($E, \Gamma$), $\Gamma$) $=$nd($E, \Gamma$) $- 1$. Let $G$ be the complete graph over all the symbols in $\mathcal{S}$. By the same argument as in lemma 5, small-edges($G, \Gamma$) $\supseteq H$, where $H$ is the connected components of the shorts of $\mathcal{S}$, as computed in reduce_constraints($\mathcal{S}$). Let $\mathcal{S}'$ be the set of constraints whose longs are in small-edges($G, \Gamma$). It follows that $\mathcal{S}' \supseteq$ reduce_constraints($\mathcal{S}$). Now small-edges($G, \Gamma$) $\supseteq$ edges($\mathcal{S}'$) $\supseteq$ edges(reduce_constraints($\mathcal{S}$)).
Hence nd(edges($\mathcal{S}$), $\Gamma$) $=$ nd($G, \Gamma$) $=$ nd(small-edges($G, \Gamma$), $\Gamma$) $+ 1$ $\geq$ nd(edges(reduce_constraints($\mathcal{S}$))) $+ 1 \geq$ (by the inductive hypothesis) num_labels(reduce_constraints($\mathcal{S}$)) $+ 1 =$ num_labels($\mathcal{S}$). $\square$

**Theorem 2:** Out of all solutions to the set of constraints $\mathcal{S}$, the instantiations of solve_constraints($\mathcal{S}$) have the fewest number of different values of od($a, b$), where $a, b$ range over the symbols in $\mathcal{S}$. This number is given by num_labels($\mathcal{S}$).

**Proof:** Immediate from lemma 18.

**Corollary 19:** Let $\Omega$ have all the properties of an om-space except that it has only $k$ different orders of magnitude. A system of constraints $\mathcal{S}$ has a solution in $\Omega$ if and only if the tree returned by solve_constraints($\mathcal{S}$) uses no more than $k$ different labels.

**Proof:** Immediate from theorems 1 and 2. □

### Proof of Algorithm for Non-strict Comparisons

We now prove that the revised algorithm presented in Section 6.2 for non-strict comparisons is correct. The proof is only a slight extension of the proof of theorem 1, given above.

Recall that the revised algorithm in Section 6.2 replaces the line of solve_constraints

$H$ := the connected components of the shorts of $\mathcal{S}$

with the following code:

```
1.    H := the shorts of S;
2.    repeat H := the connected components of H;
3.           for each weak constraint od(a, b) ≪ od(c, d)
4.                if cd is in H then add ab to H endif endfor
5.    until no change has been made to H in the last iteration.
```

We need the following new lemmas and proofs:

**Lemma 20:** Let $\mathcal{S}$ be a set of strict comparisons, and let $\mathcal{W}$ be a set of non-strict comparisons. Let $H$ be the set of edges output by the above code. If $\mathcal{S} \cup \mathcal{W}$ is consistent, then there is an edge in $\mathcal{S}$ that is not in $H$.

**Proof:** As in the proof of lemma 5, let $\Gamma$ be a valuation satisfying $\mathcal{S} \cup \mathcal{W}$ and let $pq$ be an edge in $\mathcal{S}$ such that od($\Gamma(p), \Gamma(q)$) is maximal. We wish to show that, for every edge $ab \in H$, od($\Gamma(a), \Gamma(b)$) $\ll$ od($\Gamma(p), \Gamma(q)$), and hence $ab \neq pq$. Proof by induction: suppose that this holds for all the edges in $H$ at some point in the code, and that $ab$ is now to be added to $H$. There are three cases to consider.

- $ab$ is added in step [1]. Then, as in lemma 5, there is a constraint od($a, b$) $\ll$ od($c, d$) in $\mathcal{S}$. Hence od($\Gamma(a), \Gamma(b)$) $\ll$ od($\Gamma(c), \Gamma(d)$) $\lll$ od($\Gamma(p), \Gamma(q)$).

- $ab$ is added in step [2]. Then there is a path $a_1 = a, a_2 \ldots a_k = b$ such that the edge $a_i a_{i+1}$ is in $H$ for $i = 1 \ldots k-1$. By the inductive hypothesis, od($\Gamma(a_i), \Gamma(a_{i+1})$) $\ll$ od($\Gamma(p), \Gamma(q)$). By the om-triangle inequality,
od($\Gamma(a), \Gamma(b)$) $\lll$ $\max_{i=1..k-1}$(od($\Gamma(a_i), \Gamma(a_{i+1})$)) $\ll$ od($\Gamma(p), \Gamma(q)$).

- $ab$ is added in step [4]. Then there is a constraint od($a, b$) $\lll$ od($c, d$) in $\mathcal{W}$ such that $cd$ is in $H$. By the inductive hypothesis, od($\Gamma(c), \Gamma(d)$) $\ll$ od($\Gamma(p), \Gamma(q)$).

☐

**Lemma 21:** Let $\mathcal{W}$ contain the constraint $\mathrm{od}(a, b) \lll \mathrm{od}(c, d)$. Suppose that the algorithm returns a cluster tree $T$. Let $M$ be the least common ancestor of $a$ and $b$ in $T$, and let $N$ be the least common ancestor of $c$ and $d$. Then $M$.label $\leq N$.label.

**Proof:** By lemma 13, $N$ is assigned a label in the first iteration where $H$ does not include the edge $cd$. In all previous iterations, since $cd$ is in $H$, $ab$ will likewise be put into $H$. Hence $M$ does not get assigned a label before $N$, so $M$.label $\leq N$.label.

The remainder of the proof of the correctness of the revised algorithm is exactly the same as the proof of theorem 1.

### Validation of Algorithm Solve_constraints2

The proof of the correctness of algorithm solve_constraints2 is again analogous in structure to the proof of theorem 1. We sketch it below: the details are not difficult to fill in.

1. (Analogue of lemma 2:) If $T$ is an ordered cluster tree, then the revised version of instantiate($T$) returns an instantiation of $T$. The proof is exactly the same as lemma 2, with the additional verification that instantiate2 preserves the orderings in $T$.

2. (Analogue of lemma 5:) Let $\mathcal{S}$ be a set of order-of-magnitude constraints on distances, and let $\mathcal{O}$ be a set of ordering constraints on points. Let $H$ be the graph given by the two statements

   $H :=$ the connected components of the shorts of $\mathcal{S}$;
   $H :=$ incorporate_order($H, \mathcal{O}$);

   If $\mathcal{S}$ and $\mathcal{O}$ are consistent, then $H$ does not contain all the edges of $\mathcal{S}$.

   **Proof:** As in the proof of lemma 5, choose a valuation $\Gamma$ satisfying $\mathcal{S}, \mathcal{O}$ and let $pq$ be an edge in $\mathcal{S}$ for which $\mathrm{od}(\Gamma(p), \Gamma(q))$ is maximal. Following the informal argument presented in Section 6.3, it is easily shown that $pq$ is longer than any of the edges added in these two statements, and hence it is not in $H$.

3. (Analogue of lemma 9:) If solve_constraints2 returns **false**, then $\mathcal{S}, \mathcal{O}$ is inconsistent.
   **Proof:** Immediate from (2).

4. (Analogue of lemma 12:) If solve_constraints2($\mathcal{S}, \mathcal{O}$) does not return false, then it returns a well-formed ordered cluster tree.

   **Proof:** By merging the strongly connected components of $G$, incorporate_order always ensures that the ordering arcs between connected components of $H$ form a DAG. These arcs are precisely the same ones that are later added among the children of node $N$ as ordering arcs. Thus, the ordering arcs over the children of a node in the cluster tree form a DAG. Otherwise, the construction of the tree $T$ is the same as in lemma 12.

The remainder of the proof is the same as the proof of theorem 1.

**Proof of Theorem 3**

We begin by proving lemma 22, that the revised version of "instantiate", given in Section 6.3, gives an instantiation of a cluster tree in Euclidean space.

**Lemma 22:** Any cluster tree $T$ has an instantiation in Euclidean space $\Re^m$ of any dimensionality $m$.

The proof is essentially the same as the proof of Lemma 2, except that we now have to keep track of real quantities. For any node $N$, if $i=N$.label, we define $\Delta(N) = \delta_i$. The proof then proceeds in the following steps:

i. For any $i < j$, $\delta_i < \delta_j / \alpha^{j-i}$. Immediate by construction.

ii. For any nodes $M, C$, if $M$ is a descendant of $C$ in $T$ then
$\text{dist}(G[M], G[C]) < \alpha n \Delta(C) / (\alpha - 1)$.

 **Proof:** Let $C = C_0, C_1 \ldots C_r = M$ be the path from $C$ to $M$ through $T$. Then $\text{dist}(G[M], G[C]) \leq$ (by the triangle inequality) $\sum_{i=0}^{r-1} \text{dist}(G[C_{i+1}], G[C_i]) \leq \sum_{i=0}^{r-1} (n\Delta(C)/\alpha^i) < (\alpha/(\alpha-1))(n\Delta(C))$.

iii. Let $N$ be a node in $T$; let $C_1$ and $C_2$ be two children of $N$; and let $M_1$ and $M_2$ be descendants of $C_1$ and $C_2$ respectively. Then
$\Delta(N)(1 - 2n/(\alpha - 1)) < \text{dist}(G[M_1], G[M_2]) < n\Delta(N)(1 + 2/(\alpha - 1))$
 **Proof:** By the triangle inequality,
$\text{dist}(G[C_1], G[C_2]) \leq \text{dist}(G[C_1], G[M_1]) + \text{dist}(G[M_1], G[M_2]) + \text{dist}(G[M_2], G[C_2])$.
Thus, $\text{dist}(G[C_1], G[C_2]) - \text{dist}(G[C_1], G[M_1]) - \text{dist}(G[M_2], G[C_2]) \leq \text{dist}(G[M_1], G[M_2])$.
Also, by the triangle inequality,
$\text{dist}(G[M_1], G[M_2]) \leq \text{dist}(G[C_1], G[C_2]) + \text{dist}(G[C_1], G[M_1]) + \text{dist}(G[M_2], G[C_2])$.
By construction, $\Delta(N) \leq \text{dist}(G[C_1], G[C_2]) < n\Delta(N)$,
and by part (ii), for $i = 1, 2$, $\text{dist}(G[M_i], G[C_i]) < \alpha n \Delta(C)/(\alpha-1) < n\Delta(N)/(\alpha-1)$
as $\Delta(C) < \Delta(N)/\alpha$.

iv. For any symbols $a, b, c, d$ in $T$, let $P$ be the least common ancestor of $a, b$ and let $N$ be the least common ancestor of $c, d$. If $P$.label $< N$.label then
much_closer($G[a], G[b], G[c], G[d]$).

 **Proof:** By part (iii), $\text{dist}(G[a], G[b]) < n\Delta(P)(1 + 2/(\alpha - 1))$
and $\text{dist}(G[c], G[d]) > \Delta(N)(1 - 2n/(\alpha - 1))$. Since $\Delta(P) < \Delta(N)/\alpha$ and since $\alpha = 2 + 2n + Bn$, it follows by straightforward algebra that
$\text{dist}(G[a], G[b]) < \text{dist}(G[c], G[d]) \ / \ B$.

$\square$

We next prove the analogue of lemma 5.

**Lemma 23:** Let $\mathcal{S}$ be a set of constraints over $n$ variables of the form
"$\text{dist}(a, b) < \text{dist}(c, d) \ / \ B$", where $B > n$. If $\mathcal{S}$ is consistent, then there is some edge in $\mathcal{S}$ which is not in the connected components of the shorts of $\mathcal{S}$.

**Proof:** Let $\Gamma$ be a valuation satisfying $\mathcal{S}$. Let $pq$ be the edge in $\mathcal{S}$ for which $\text{dist}(\Gamma(p), \Gamma(q))$ is maximal. Now, if $ab$ is a short of $\mathcal{S}$ — that is, there is a constraint much_closer($a, b, c, d$) in $\mathcal{S}$ — then $\text{dist}(\Gamma(a), \Gamma(b)) < \text{dist}(\Gamma(c), \Gamma(d))/B \leq \text{dist}(\Gamma(p), \Gamma(q))/B$.

Now, let $ab$ be any edge in $H$, the connected components of the shorts of $\mathcal{S}$. Then there is a simple path $a_1 = a, a_2 \ldots a_k = b$ such that the edge $a_i a_{i+1}$ is a short of $\mathcal{S}$ for $i = 1 \ldots k - 1$. Note that $k \leq n$. Then, by the triangle inequality,

$\text{dist}(\Gamma(a), \Gamma(b)) \leq$
$\text{dist}(\Gamma(a_1), \Gamma(a_2)) + \text{dist}(\Gamma(a_2), \Gamma(a_3)) + \ldots + \text{dist}(\Gamma(a_{k-1}), \Gamma(a_k)) \leq$
$(k-1)\text{dist}(\Gamma(p), \Gamma(q)) \, / \, B < \text{dist}(\Gamma(p), \Gamma(q))$

Hence $pq \neq ab$, so $pq$ is not in $H$. $\square$

**Theorem 3:** Let $\mathcal{S}$ be a set of constraints over $n$ variables of the form "$\text{dist}(a, b) < \text{dist}(c, d)$ $/$ $B$", where $B > n$. The algorithm solve_constraints($\mathcal{S}$) returns a cluster tree satisfying $\mathcal{S}$ if $\mathcal{S}$ is consistent over Euclidean space, and returns **false** if $\mathcal{S}$ is inconsistent.

**Proof:** Note that the semantics of the constraints "much_closer$(a, b, c, d)$" enters into the proof of Theorem 1 only in lemmas 2 and 5. The remainder of the proof of Theorem 1 has to do purely with the relation between the structure of $\mathcal{S}$ and the structure of the tree. Hence, since we have shown that the analogues of lemmas 2 and 5 hold in a set of constraints of this kind, the same proof can be completed in exactly the same way. $\square$

**Proof of Theorem 4**

**Lemma 24:** Let $T$ be a cluster tree and let $\Gamma$ be a valuation over om-space $\Omega$ satisfying $T$. Let $x$ be a symbol not in $T$, let $a$ be a point in $\Omega$, and let $\Gamma'$ be the valuation $\Gamma \cup \{x \to a\}$. Then there exists an extension $T'$ of $T$ by $x$ such that $\Gamma'$ satisfies $T'$.

**Proof:** If $T$ is the empty tree, the statement is trivial. If $T$ contains the single symbol $y$, then if $a = \Gamma(y)$ then operation (2) applies with $M$.label=0; if $a \neq \Gamma(y)$ then operation (2) applies with $M$.label=1.

Otherwise, let $y$ be the symbol in $T$ such that $\text{od}(\Gamma(y), a)$ is minimal. (We will deal with the case of ties in step (D) below.) Let $F$ be the father of $y$ in $T$.

Let $D = \text{od}(\Gamma(y), a)$. Let $V$ be the set of all orders of magnitude of $\text{od}(\Gamma(p), \Gamma(q))$, where $p$ and $q$ range over symbols in $T$. We define $L$ to be the *suitable label for $D$* as follows: If $D \in V$, then $L$ is the label in $T$ corresponding to $D$. If $D$ is larger than any value in $V$ then $L$ is the label of the root of $T$ plus 1. If $D \notin V$, but some value in $V$ is larger than $D$, then let $D_1$ be the largest value in $V$ less than $D$; let $D_2$ be the smallest value in $V$ greater than $D$; let $L_1$, $L_2$ be the labels in $T$ corresponding to $D_1$, $D_2$; and let $L = (L_1 + L_2)/2$.

One of the following must hold:

A. $\Gamma(y) = a$, and $F$.label=0. Then apply operation (3) with $N = F$.

B. $\Gamma(y) = a$ and $F$.label $\neq 0$. Then apply operation (4) with $M$.label = 0.

C. $\Gamma(y) \neq a$, but $\text{od}(\Gamma(y), a)$ is less than $\text{od}(\Gamma(z), a)$ for any other symbol $z \neq y$ in $T$. Apply operation (4) with $M$.label set to the suitable value for $D$ in $T$.

D. There is more than one value $y_1 \ldots y_k$ for which $\text{od}(\Gamma(y_i), a) = D$. It is easily shown that in this case there is an internal node $Q$ such that $y_1 \ldots y_k$ is just the set of symbols in the subtree of $Q$. There are three cases to consider:

D.i $D=$odiam($\Gamma(Q$.symbols)). Then apply operation (3) with $N = Q$.

D.ii $D >$ odiam($\Gamma(Q$.symbols)), and $Q$ is not the root. Then apply operation (5) with $C = Q$. Set $M$.label to be the suitable value for $D$. (It is easily shown that $D <$ odiam($\Gamma(N$.symbols)), where $N$ is the father of $Q$.)

D.iii $D >$ odiam($\Gamma(Q$.symbols)), and $Q$ is the root. Apply operation (6).

□

**Lemma 25:** Let $A = \{a_1 \ldots a_k\}$ be a finite set of points whose diameter has order-of-magnitude $D$. Then there exists a point $u$ such that, for $i = 1 \ldots k$, od$(u, a_i) = D$.

**Proof:** Let $b_1 = a_1$. By axiom A.8 there exists an infinite collection of points $b_2, b_3 \ldots$ such that od$(b_i, b_j) = D$ for $i \neq j$. Now, for any value $a_i$ there can be at most one value $b_j$ such that od$(a_i, b_j) \ll D$; if there were two such values $b_{j1}$ and $b_{j2}$, then by the om-triangle inequality, od$(b_{j1}, b_{j2}) \ll D$. Hence, all but $k$ different values of $b_j$ are at least $D$ from any of the $a_i$. Let $u$ be any of these values of $b_j$. Then since od$(u, a_1) = D$ and od$(a_1, a_i) \lll D$ for all $i$, it follows that od$(u, a_i) \lll D$ for all $a_i$. Thus, since od$(u, a_i) \lll D$ but not od$(u, a_i) \ll D$, it follows that od$(u, a_i) = D$. □

**Lemma 26:** Let $T$ be a cluster tree; let $\Gamma$ be a valuation over om-space $\Omega$ satisfying $T$; and let $T'$ be an extension of $T$ by $x$. If $\Omega$ is dense and unbounded above, then there is a value $a$ such that the valuation $\Gamma \cup \{x \rightarrow a\}$ satisfies $T'$.

**Proof:** For operations (1) and (2) the statement is trivial.

Otherwise, let $L$ be an extending label of $T$. If $L = 0$, then set $D = 0$. If $L$ is in $T$, then let $D$ be the order of magnitude corresponding to $L$ in $T$ under $\Gamma$. If $L_1 < L < L_2$ where $L_1$ and $L_2$ are labels of consecutive values in $T$, then let $D_1$ and $D_2$ be the orders of magnitude corresponding to $L_1, L_2$ in $T$ under $\Gamma$. Let $D$ be chosen so that $D_1 \ll D \ll D_2$. If $L$ is greater than any label in the tree, then choose $D$ to be greater than the diameter of the tree under $\Gamma$.

If $T'$ is formed from $T$ by operation (3), then using lemma 25 let $a$ be a point such that od$(a, \Gamma(y)) = $ odiam$(N)$ for all $y$ in $N$.symbols.

If $T'$ is formed from $T$ by operation (4), then let $a$ be a point such that od$(a, \Gamma(y)) = D$.

If $T'$ is formed from $T$ by operation (5), then let $a$ be a point such that od$(a, \Gamma(y)) = D$ for all $y$ in $C$.symbols. (Note that, since $M$.label $< N$.label, $D <$ odiam$(N$.symbols).)

If $T'$ is formed from $T$ by operation (6), then let $a$ be a point such that od$(a, \Gamma(y)) = D$ for all $y$ in $R$.symbols.

In each of these cases, it is straightforward to verify that $\Gamma \cup \{x \rightarrow a\}$ satisfies $T'$. □

As we observed in Section 8 regarding lemma 28, the conditions on $\Omega$ in lemma 26 are necessary, and the statement is false otherwise. For example, let $\Omega$ be the om-space described in example I, Section 3, of polynomials over an infinitesimal $\delta$. Then $\Omega$ is not unbounded above; there is a maximum order-of-magnitude $O(1)$. Let $T$ be the starting tree of Figure 3 (upper-left corner), and let $T'$ be the result of applying operation 6 (middle bottom). Let $\Gamma$ be the valuation $\{u \rightarrow \delta, v \rightarrow 2\delta, w \rightarrow 1\}$. Then $\Gamma$ satisfies $T$, but it cannot be extended to a valuation that satisfies $T'$, as that would require $x$ to be given a value such that od$(v, w) \ll$ od$(x, w)$, and no such value exists within $\Omega$. The point of the lemma is that, if $\Omega$ is required to be both dense and unbounded above, then we cannot get "stuck" in this way.

**Lemma 27:** Let $T$ be a cluster tree. Let $X$ be a variable not among the symbols of $T$. Let $\alpha$ be an open formula in $\mathcal{L}$, whose free variables are the symbols of $T$ and the variable $X$. Let $\phi$ be the formula $\exists_X \alpha$. Let $\Omega$ be an om-space that is dense and unbounded above. Then there exists an instantiation $\Gamma$ of $T$ in $\Omega$ that satisfies $\phi$ if and only if there exists an extension $T'$ of $T$ and an instantiation $\Gamma'$ of $T'$ that extends $\Gamma$ and satisfies $\alpha$.

**Proof:** Suppose that there exists an instantiation $\Gamma$ of $T$ that satisfies $\exists_X \alpha$. Then, by definition, there is a point $a$ in $\Omega$ such that $\Gamma$ satisfies $\alpha(X/a)$. That is, the instantiation $\Gamma \cup \{X \rightarrow a\}$ satisfies $\alpha$. Let $\Gamma' = \Gamma \cup \{X \rightarrow a\}$. By lemma 24, the cluster tree $T'$ corresponding to $\Gamma'$ is an extension of $T$.

Conversely, suppose that there exists an extension $T'$ of $T$ and an instantiation $\Gamma'$ of $T'$ satisfying $\alpha$. Let $\Gamma$ be the restriction of $\Gamma'$ to the symbols of $T$. Then clearly $\Gamma$ satisfies the formula $\exists_X \alpha$. $\square$

**Lemma 28:** Let $T$ be a cluster tree. Let $\phi$ be an open formula in $\mathcal{L}$, whose free variables are the symbols of $T$. Let $\Omega$ be an om-space that is dense and unbounded above. If one instantiation $\Gamma$ of $T$ in $\Omega$ satisfies $\phi$ then every instantiation of $T$ in $\Omega$ satisfies $\phi$.

**Proof:** We can assume without loss of generality that the only logical symbols in $\phi$ are $\neg$ (not), $\wedge$ (and), $\exists$ (exists), $=$ (equals) and variables names, and that the only non-logical symbol is the predicate "much_closer". We now proceed using structural induction on the form of $\phi$. Note that an equivalent statement of the inductive hypothesis is, "For any formula $\psi$, either $\psi$ is true under every instantiation of $T$, or $\psi$ is false under every instantiation of $T$."

Base case: If $\phi$ is an atomic formula "$X = Y$" or "much_closer$(W, X, Y, Z)$" then this follows immediately from corollary 4.

Let $\phi$ have the form $\neg \psi$. If $\phi$ is true under $\Gamma$, then $\psi$ is false under $\Gamma$. By the inductive hypothesis, $\psi$ is false under every instantiation of $T$. Hence $\phi$ is true under every instantiation of $T$.

Let $\phi$ have the form $\psi \wedge \theta$. If $\phi$ is true under $\Gamma$ then both $\psi$ and $\theta$ are true under $\Gamma$. By the inductive hypothesis, both $\psi$ and $\theta$ are true under every instantiation of $T$. Hence $\phi$ is true under every instantiation of $T$.

Let $\phi$ have the form $\exists_X \alpha$. If $\phi$ is true under $\Gamma$ then by lemma 27, there exists an extension $T'$ of $T$ and a instantiation $\Gamma'$ of $T'$ such that $\alpha$ is true under $\Gamma'$. By the inductive hypothesis, $\alpha$ is true under every instantiation of $T'$. Now, if $\Delta'$ is an instantiation of $T'$ that satisfies $\alpha$, and $\Delta$ is the restriction of $\Delta'$ to the variables in $T$, then clearly $\Delta$ satisfies $\exists_X \alpha$. But by lemma 26, every instantiation of $T$ can be extended to an instantiation $\Delta'$ of $T'$. Therefore, every instantiation of $T$ satisfies $\phi$. $\square$

**Theorem 4:** Let $T$ be a cluster tree. Let $\phi$ be an open formula in $\mathcal{L}$, whose free variables are the symbols of $T$. Let $\Omega$ be an om-space that is dense and unbounded above. Algorithm decide$(T, \phi)$ returns **true** if $T$ satisfies $\phi$ and **false** otherwise.

**Proof:** Immediate from the proof of lemma 28. $\square$

Davis

## References

Cormen, T.H., Leiserson, C.E., and Rivest. R.L. (1990). *Introduction to Algorithms.* Cambridge, MA: MIT Press

Davis, E. (1990). Order of Magnitude Reasoning in Qualitative Differential Equations. In D. Weld and J. de Kleer (Eds.) *Readings in Qualitative Reasoning about Physical Systems.* San Mateo, CA: Morgan Kaufmann. 422-434.

Keisler, J. (1976). *Foundations of Infinitesimal Calculus.* Boston, MA: Prindle, Webber, and Schmidt.

Mavrovouniotis, M. and Stephanopoulos, G. (1990). "Formal Order-of-Magnitude Reasoning in Process Engineering." In D. Weld and J. de Kleer (Eds.) *Readings in Qualitative Reasoning about Physical Systems.* San Mateo, CA: Morgan Kaufmann. 323-336.

Raiman, O. (1990). "Order of Magnitude Reasoning." In D. Weld and J. de Kleer (Eds.) *Readings in Qualitative Reasoning about Physical Systems.* San Mateo, CA: Morgan Kaufmann. 318-322.

Robinson, A. (1965). *Non-Standard Analysis.* Amsterdam: North-Holland Publishing Co.

Weld, D. (1990). "Exaggeration." In D. Weld and J. de Kleer (Eds.) *Readings in Qualitative Reasoning about Physical Systems.* San Mateo, CA: Morgan Kaufmann. 417-421.

38